\RequirePackage{fix-cm}
\documentclass[smallextended]{svjour3}       
\smartqed  
\usepackage[hyphens]{url}  
\usepackage{amsmath,amssymb,amsfonts} 
\usepackage{booktabs} 
\usepackage{caption}  
\usepackage{subcaption} 
\usepackage{graphicx}
\usepackage{natbib} 

\usepackage[usenames,dvipsnames]{color}
\usepackage{colortbl}
\definecolor{mygray}{gray}{.9}

\begin{document}
\title{A Spatiotemporal Deep Neural Network for Fine-Grained Multi-Horizon Wind Prediction  
}

\titlerunning{Multi-Horizon SpatioTemporal Network}        

\author{Fanling Huang         \and
        Yangdong Deng 
}


\institute{Fanling Huang \at
              School of Software, Tsinghua University, Beijing, China \\
              \email{fanlinghuang@163.com}           
           \and
           Yangdong Deng \at
              School of Software, Tsinghua University, Beijing, China \\
              \email{dengyd@mail.tsinghua.edu.cn}
}

\date{Received: date / Accepted: date}

\maketitle
\begin{abstract}

The prediction of wind in terms of both wind speed and direction, which has a crucial impact on many real-world applications like aviation and wind power generation, is extremely challenging due to the high stochasticity and complicated correlation in the weather data. Existing methods typically focus on a sub-set of influential factors and thus lack a systematic treatment of the problem. In addition, fine-grained forecasting is essential for efficient industry operations, but has been less attended in the literature. In this work, we propose a novel data-driven model, Multi-Horizon SpatioTemporal Network (MHSTN), generally for accurate and efficient fine-grained wind prediction. MHSTN integrates multiple deep neural networks targeting different factors in a sequence-to-sequence (Seq2Seq) backbone to effectively extract features from various data sources and produce multi-horizon predictions for all sites within a given region. MHSTN is composed of four major modules. First, a temporal module fuses coarse-grained forecasts derived by Numerical Weather Prediction (NWP) and historical on-site observation data at stations so as to leverage both global and local atmospheric information. Second, a spatial module exploits spatial correlation by modeling the joint representation of all stations. Third, an ensemble module weighs the above two modules for final predictions. Furthermore, a covariate selection module automatically choose influential meteorological variables as initial input. MHSTN is already integrated into the scheduling platform of one of the busiest international airports of China. The evaluation results demonstrate that our model outperforms competitors by a significant margin.

\keywords{Wind forecast  \and Time series prediction \and Neural networks \and Spatiotemporal data mining}
\end{abstract}

\begin{figure}[htbp]
\centering
\includegraphics[width=0.6\textwidth]{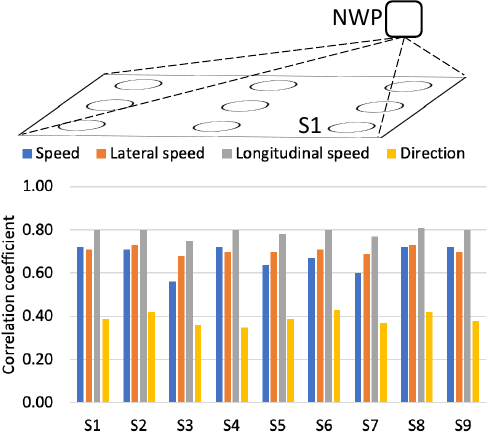}
\caption{Top: The schema of an airfield with nine local weather observation stations for fine-grained monitoring of wind and a global NWP service offering coarse-grained forecasting over the whole area. Bottom: The correlations between the local observations and the NWP predictions are generally significant but far from perfect (1.00). The detailed amount of correlation varies across locations and variables.}
\label{fig:corr-nwp}
\end{figure} 
\begin{figure}
\centering
\includegraphics[width=0.95\textwidth]{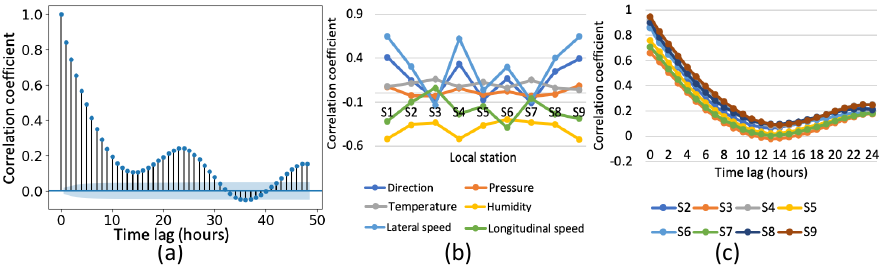}
\caption{Weather data exhibit complicated correlations. (a), auto correlations of wind speed observed at station $S1$; (b), cross correlations between wind speed and other weather variables at stations; (c), spatial correlations of wind speed between the station $S1$ and the other stations changing along different time lags. }
\label{fig:corr-obs}
\end{figure}
\section{Introduction}

Wind, which is an essential natural phenomenon determined by both wind speed and wind direction, plays a vital role in human life. For example, it is essential to forecast wind speed and wind direction for safe and efficient seafaring and flight \footnote{\url{https://www.aviationweather.gov/windtemp}}.
However, wind has been widely recognized as one of the most difficult meteorological attributes to forecast due to its high stochasticity and complicated correlation schema \citep{bauer2015the,wilson2018a,masseran2013fitting}.
In this work, we focus on forecasting both wind speed and wind direction at a finer granularity (e.g., a 1 km by 1 km grid) over a relatively long period of time (e.g., up to 24 hours) by leveraging both global Numerical Weather Prediction (NWP) and local observation data. Fine-grained prediction is of great importance in many real-world applications. For example, a recent study \citep{ezzat2020turbine} claimed that \textit{"the operation of today’s massive-scale turbines has urged the need for forecasts of higher spatial resolution than those conventionally issued at the farm level."}
This work is presented through the lens of a wind forecast application deployed in the scheduling platform of a leading international airport (Fig.~\ref{fig:corr-nwp}), which has a strong demand to improve the latency and precision of wind prediction for a higher level of operation efficiency. 
Our work enables a systematic treatment of a series of previously less attended problems, which turn out to be critical in boosting prediction performance.

\textbf{Problem 1: How to fuse global NWP and local observation data for accurate and timely fine-grained prediction over a relatively long period of time?}
NWP, which numerically solves atmospheric equations with high-performance computers, has been made available at a global scale \citep{bauer2015the}. However, it is too expensive even infeasible for fine-grained predictions, particularly at a high-spatial resolution. In our case (Fig.~\ref{fig:corr-nwp}), all stations of the airfield share a single source of NWP.
In contrast, statistical models, especially deep neural networks (DNNs) \citep{zhang2019long,qiong2019direct}, learning from local observation data, have shown great potential for fast forecasting. Nonetheless, such methods are less effective in forecasting long time horizons than NWP does. 
The empirical results in Fig.~\ref{fig:corr-nwp} suggest that NWP outputs are valuable for each station to make fine-grained forecasts by involving variances inherent in local observation data. 

\textbf{Problem 2: How to leverage complicated correlation among various factors?}
Weather data observed across multiple stations at a local field exhibit complicated correlation patterns as shown in Fig.~\ref{fig:corr-obs}. The observation of a variable at a time is correlated with its historical behaviors (auto-correlation), substantially influenced by other variables (cross-correlation), and also associated with neighboring observations (spatial correlation).
Therefore, assimilating the above correlations is potentially highly valuable to boost the prediction accuracy.
Such correlations, nevertheless, are just partially addressed in existing works \citep{zhang2019long,wilson2018a}.
In addition, compared with previous studies that focus on global- or regional- level prediction, forecasting wind at a high-spatial resolution suffers from significant fluctuations and dependencies \citep{ezzat2020turbine}.

\textbf{Problem 3: How to derive multi-horizon predictions?}
Multi-horizon prediction, i.e., forecasting on multiple steps in future time, is preferred in many practical circumstances as it offers guidance for operation scheduling over a period of time. However, the majority of statistical wind prediction models are limited to one-step ahead forecasting \citep{wang2016an,grover2015a,wilson2018a}. Recent survey papers \citep{wang2016an,wen2017a} recommended a direct multi-horizon strategy as it can avoid error accumulation, capture the dependencies between predictions, and retain efficiency by parameter sharing (please refer to section~\ref{sec:rel} for more details). 

\textbf{Problem 4: How to predict wind direction?} 
It is essential for scheduling airport operations ahead of time. For instance, an airport can change the direction of take off when the wind direction is significantly changed. 
Although wind direction prediction is always included in NWP, it is far less addressed by machine learning methods \citep{masseran2013fitting,erdem2011arma}, perhaps because wind direction is harder to be analyzed and modeled. Especially, wind direction is a periodic variable and thus the conventional distributions are infeasible to model it. The hardness could also be illustrated by the low correlation values of wind direction in Fig.~\ref{fig:corr-nwp}. Therefore, it is meaningful to model wind direction in a statistical manner for efficient prediction.

To address the above problems, we propose a generic framework, dubbed as Multi-Horizon SpatioTemporal Network (MHSTN), that produce multi-horizon predictions at a high-spatial resolution for both wind speed and wind direction. As a systematic effort to boost wind prediction for industry-level applications, MHSTN takes a sequence-to-sequence (Seq2Seq) structure as backbone \citep{sutskever2014sequence} that naturally implements the direct multi-horizon strategy as well as in which advanced deep learning techniques can be seamlessly utilized. We integrate four major modules into MHSTN to predict a target variable of all stations across a local area, First, we devise a temporal module by hybridizing Long Short-Term Memory Network (LSTM) and MultiLayer Perceptron (MLP) to combine both locally historical and globally NWP temporal data. Second, we employ a Convolutional Neural Network (CNN) as a spatial module to constitute a complete representation of the region base on the state representations of respective stations. Third, we weigh the above two modules with an ensemble module to produce final predictions for each station. Further, we devise a covariate selection module to pick up more influential meteorological variables as the initial input. Our primary contributions are as follows:
\begin{itemize}
    \item We propose a novel data-driven model, MHSTN, for accurate and timely fine-grained wind prediction. To our knowledge, it is the first one to treat all the aforementioned challenges in a unified framework.
    \item We evaluate MHSTN with a real-world dataset collected from one of the busiest airports in the whole world. Our models are efficient and significantly outperform the competitor models in terms of prediction accuracy. In comparison with the on-site NWP results, the average reductions in prediction error are up to 13.59\% and 5.48\% for wind speed and wind direction, respectively. It should be noted that our design does not depend on the particular airfield, so MHSTN is generally applicable to any other region.
    \item We publicize our dataset and source code \footnote{\url{https://github.com/hfl15/windpred.git}} for future research. To our knowledge, it is the first open project involving multi-station observation data and NWP data for fine-grained wind prediction. It is also the first targeting to an airport. 
\end{itemize}

The rest of the paper is organized as follows. 
We first discuss related work in section \ref{sec:rel}. 
In section \ref{sec:prob}, we formulate the problem. 
In section \ref{sec:app}, we introduce the design details of our integrated framework as well as its major building blocks. 
In section \ref{sec:exp}, we report the real-world data used in this work and present the comprehensive experiments to evaluate our framework.
Finally, we conclude the paper and outline future work in section \ref{sec:con}.

\section{Related Work}
\label{sec:rel}

\subsection{Wind prediction}

Wind prediction as a core task in NWP has been obtained substantial improvement in recent decades \citep{bauer2015the}. However, NWP models are limited to coarse-grained predictions due to the excessive demand for computing power. 
In contrast, some studies have demonstrated that statistical models \citep{wang2016an,cavalcante2017lasso}, especially the DNN models \citep{qiong2019direct,zhang2019long,zhang2018wind}, are significantly more efficient to capture regional variances. Nonetheless, these models are limited to (very) short-term forecasting due to the shortage of global atmospheric information. 
Few attempts have been taken to couple the above both sides for regional medium-/long-term wind forecasting \cite{salcedo2009hybridizing,cheng2017short}, wind power production forecasting \cite{corizzo2021multi,gonccalves2021forecasting} and weather forecasting \cite{wang2019deep}.
Along this line, we propose a new machine learning model to assimilate global NWP and local observation data and produce predictions in a higher spatial resolution.

Wind suffers from a complicated dynamic correlation schema that is often the compound of auto-correlation, cross-correlation and spatial correlation. Most existing works, nevertheless, just consider partial correlations.
First, capturing auto-correlation has been a basic capability of almost all time series models (e.g., LSTM).
Second, some studies utilized cross-correlation by leveraging handcrafted or auto-selected covariates. \cite{erdem2011arma,grover2015a} used a manually selected covariate set to boost the prediction accuracy. \cite{qiong2019direct,zhang2019long} proposed several correlation metrics to evaluate the importance of weather variables so that a set of influential covariates can be automatically selected.
Third, spatial dependence could be mined when more than one observation stations are available. \cite{damousis2004wind} encoded how the spatial correlation varies in different conditions of wind speed and wind direction via a fuzzy model into a series of rules for wind speed prediction. \cite{grover2015a} trained a predictor for each location and then combined them via a restricted Boltzman machine to form the joint distribution to derive predictions. \cite{wilson2018a} integrated graph convolutions into a LSTM network to capture the spatial correlation for wind speed prediction. \cite{khodayar2019spatiotemporal} proposed a graph deep learning model to distill the spatiotemporal features in neighboring wind farms. \cite{ceci2019spatial} focused on renewable energy production and proposed an entropy based cost function for artificial neural network framework to capture spatial information from spatially-located plants. 
Instead, we mine and fuse all of the above dynamic correlations in a unified framework.

Most existing works in wind or weather prediction are limited to one-horizon forecasting \citep{wang2016an,grover2015a,wilson2018a}. 
There are three common strategies to realize multi-horizon forecasting. (1) \textbf{Recursive strategy} trains a model to predict the one-step-ahead estimate and then iteratively feed this estimate back as the ground truth to forecast longer horizons. Due to the discrepancy between consuming actual data versus estimates during prediction, accumulated errors are usually unavoidable and enlarged by longer horizons. (2) \textbf{Direct strategy}, which separately trains multiple models with each directly predicts a future horizon, is less biased. But such a way needs significantly more computing and storage resources as well as neglects dependencies in the time series. (3) \textbf{Direct multi-horizon strategy}, which directly trains a model with a multivariate target so as to produce multi-horizon predictions simultaneously, is recommended by recent surveys \citep{taieb2015bias,wang2016an} to address the previous problems. 
Therefore, our framework use a Seq2Seq deep neural network \citep{sutskever2014sequence} as backbone to naturally implement the direct multi-horizon strategy as well as seamlessly leverage multiple advanced deep learning techniques. 

Wind is generally represented by the tuple of wind speed and wind direction \citep{erdem2011arma,grover2015a}. However, wind direction is harder to model with statistical models and thus significantly less attended in previous literatures \citep{masseran2013fitting,erdem2011arma}. 
Our framework can efficiently produce predictions for both wind speed and wind direction. 

To sum up, we propose an effective unified neural network that can be viewed as the first step towards addressing the above problems in a systematic fashion. 

\subsection{Deep neural networks}
Deep neural networks (DNNs) have been remarkably successful in many scenarios \citep{lecun2015deep}. 
MLPs are quintessential deep learning models to form the basis of many applications. 
Designed to model grid-like data, CNNs have brought about breakthroughs in the field of computer vision \citep{krizhevsky2012imagenet,karpathy2014largescale}.
Recurrent Neural Networks (RNNs) and LSTMs have been widely used to model sequential data such as speech \citep{graves2013speech} and language \citep{sutskever2014sequence}. Particularly, Seq2Seq learning is intimately related to multi-horizon time series forecasting \citep{fan2019multihorizon,wen2017a}.

Applications of DNNs to wind/weather forecasting, however, are still limited.
Typical MLPs have been introduced to cooperate with NWP models, for example \cite{rasp2018deep,salcedo2009hybridizing} used MLPs to replace parts of NWP components to accelerate the numerical simulation process, and \cite{krasnopolsky2006complex} fed the NWP predictions into a MLP for downscaling. 
\cite{qiong2019direct,zhang2019long} adapted a LSTM network to capture the history information in sequence of wind speed measurements. 
\cite{wang2019deep} proposed a Seq2Seq network with layers of RNNs to incorporate uncertainty quantification for wind speed prediction. 
Few works hybridized CNNs and RNNs/LSTMs to capture spatiotemporal features. \cite{wilson2018a} integrated graph convolutions into a LSTM network for wind speed prediction. \cite{shi2015convolutional} propose a convolutional LSTM Network for precipitation nowcasting.
Different from the above works, we take advantage of different kinds of DNNs to form a generic framework to assimilate multi-source data and thus capture complicated natures in atmosphere for both wind speed and wind direction prediction. 

\section{Problem Formulation}
\label{sec:prob}

\begin{figure}[htbp]
\centering
\includegraphics[width=0.7\textwidth]{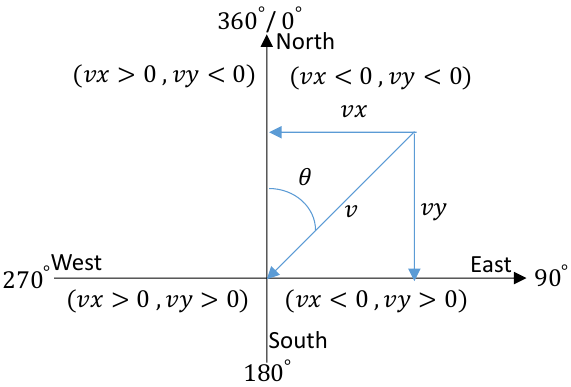}
\caption{Relation between wind speed and direction.}
\label{fig:decomp}
\end{figure}

We aim at forecasting wind speed, $v>0$, and wind direction, $\theta \in [1, 360]$, over all stations in a locally spatial field (Fig.~\ref{fig:corr-nwp}).
Note that wind direction is a periodic variable and thus conventional distributions are infeasible to model it.
To address the problem, we adopt a simple but effective strategy to infer wind direction from components of wind speed.
As shown in Fig.~\ref{fig:decomp}, wind speed ($v$) can be decomposed into a lateral component ($vx \in \mathbb{R}$) and a longitudinal component ($vy \in \mathbb{R}$) along wind direction ($\theta$). Formally:
\begin{equation}
\begin{aligned}
\label{eq:decomp-spd}
vx &= -v*sin(\frac{\theta}{360}*2\pi), \\ 
vy &= -v*cos(\frac{\theta}{360}*2\pi). \\ 
\end{aligned}
\end{equation}
Both $vx$ and $vy$ are linear variables and the respective observations are highly correlated with NWP predictions (Fig.~\ref{fig:corr-nwp}). Therefore, we turn to directly predict $vx$ and $vy$, and then the prediction of $\theta$ can be efficiently derived as follows: 
\begin{equation}
\begin{aligned}
\label{eq:spd2degree}
\theta' &= arctan(\frac{vx}{vy})/\pi*180 , \\
\theta  &=\begin{cases}
\theta' & \text{ if } vx<0\ and\ vy<0, \\
\theta' + 360 & \text{ if } vx\geqslant 0\ and\ vy<0, \\
\theta' + 180 & \text{ otherwise }.
\end{cases}
\end{aligned}
\end{equation}

Next, we introduce the problem formulation for the prediction of a wind variable ($v$, $vx$ or $vy$) over $V$ stations.
We denote the observation data as follows. 
Matrix $\mathbf{Y} \in \mathbb{R}^{T \times V}$ encapsulates the observation values of the target variable at $V$ stations over $T$ equidistant timestamps, and matrix $\mathbf{X} \in \mathbb{R}^{T \times D \times V}$ stores the values for the rest of $D$ weather variables (covariates).
Similarly, we use $\mathbf{Y}^{(f)} \in \mathbb{R}^{T \times V}$ and $\mathbf{X}^{(f)} \in \mathbb{R}^{T \times D^{(f)} \times V}$ to denote the future data (i.e., NWP data).
Both observation and NWP data are aligned to share a common timeline with the exception that the latter can approximately look into a number of future horizons.
The prediction task can be formulated as:
\begin{equation}
\label{eq:goal}
[\hat{\mathbf{Y}}_{t+1}; ...; \hat{\mathbf{Y}}_{t+K}] = F(\mathbf{Y}_{[:t]}, \mathbf{X}_{[:t]}, \mathbf{Y}^{(f)}_{[:(t+K)]},\mathbf{X}^{(f)}_{[:(t+K)]}; \mathbf{W}).
\end{equation}
The devised parametric model $F$ adaptively learns from the historical data collected from all stations and the estimated future data derived by NWP to produce $K$-horizon predictions of the target at any forecast creation time (FCT) $t$. A sequence of predictions, $\hat{\mathbf{Y}} \in \mathbb{R}^{T \times V}$, can be obtained by iterating $t$ through all FCTs that are a series of timestamps with an interval of $K$. The model parameters, $\mathbf{W}$, are trained to minimize the mean square error (MSE) defined as: 
\begin{equation}
\label{eq:loss}
MSE(\mathbf{Y}, \hat{\mathbf{Y}}) = \frac{1}{T*V}\sum_{t=1}^{T}\sum_{i=1}^{V}(\mathbf{Y}_{t,i}-\hat{\mathbf{Y}}_{t,i})^{2}.
\end{equation}
For the sake of readability, we simplify the variables related to a station $i$ as follows: $\mathbf{Y}_{.,i}$ is denoted $\mathbf{y} = [y_{1}; y_{2}; ...; y_{T}] \in \mathbb{R}^{T}$, $X_{.,.,i}$ is denoted $\mathbf{x} = [\mathbf{x}_{1}; \mathbf{x}_{2}; ...; \mathbf{x}_{T}] \in \mathbb{R}^{T \times D} $, and $\mathbf{y}^{(f)}$ and $\mathbf{x}^{(f)}$ are defined similarly. 

\section{A Unified Prediction Model}
\label{sec:app}
\begin{figure*}[htbp]
\centering
\includegraphics[width=1.0\textwidth]{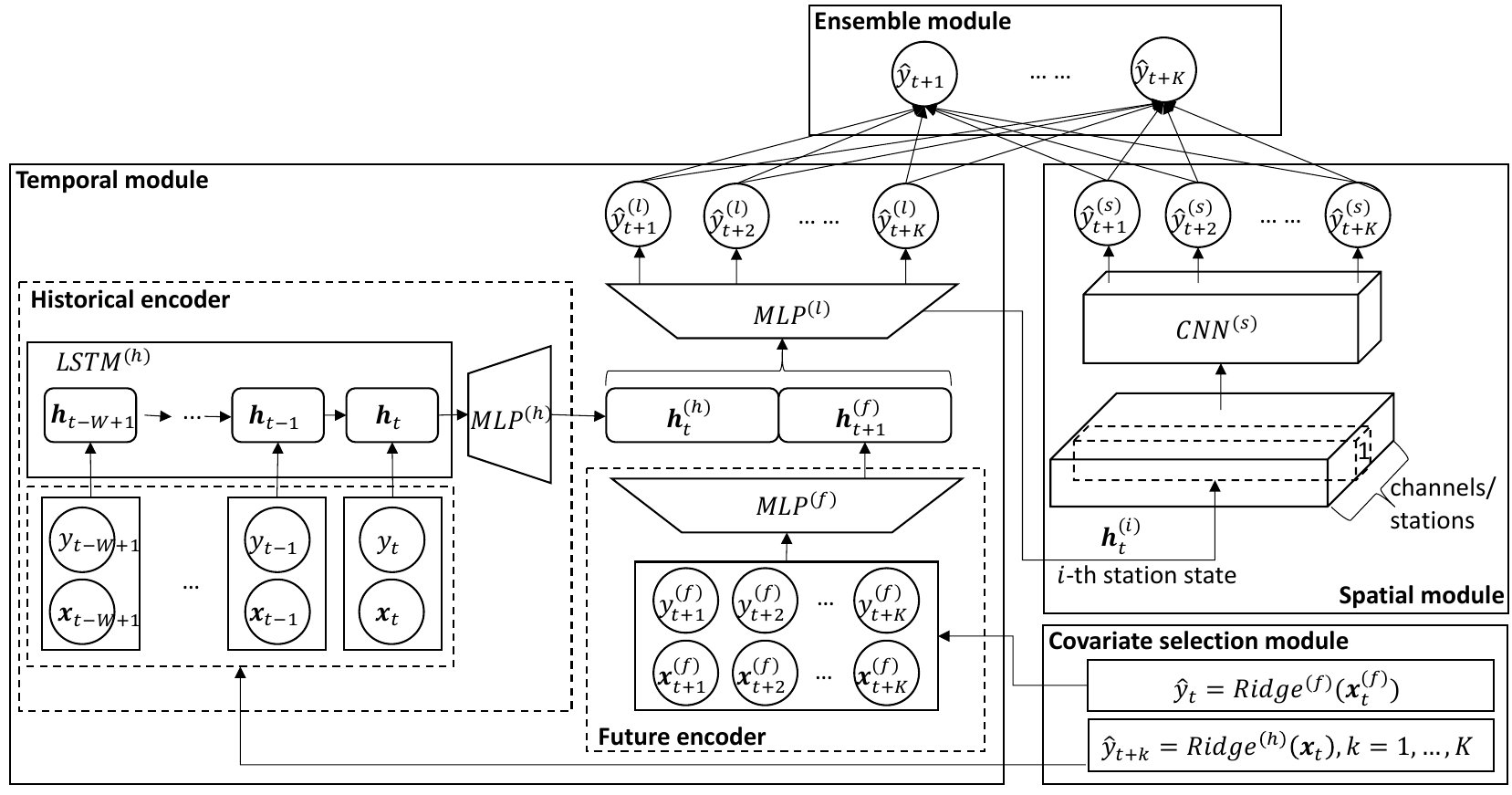}
\caption{An instance of MHSTN targeting a predicted variable (e.g., wind speed) at a station (e.g., S1).}
\label{fig:framework}
\end{figure*}

We propose a novel MHSTN model, which assimilates local observation and global NWP data, to predict wind variables ($v$, $vx$ or $vy$) of all stations in a region.
MHSTN seamlessly hybrids types of DNNs in a Seq2Seq backbone. 
As illustrated in Fig.~\ref{fig:framework}, MHSTN synergizes four major modules to predict a target variable at each station: 
\begin{enumerate}
    \item A temporal module that uses two encoders, i.e., historical encoder and future encoder, to capture the station's actual history and approximated future, respectively, and then fuses both information to generate a sequence of predictions.
    \item A spatial module that models the join of all representations of respective stations to derive a sequence of predictions for the current station.
    \item An ensemble module that weighs the above two prediction sequences to produce final output.
    \item A covariate selection module that automatically picks up influential meteorological variables as initial input to boost the above procedures.
\end{enumerate}
We will elaborate each module in following subsections.

\subsection{Temporal module}

A temporal module is trained for each station. The temporal module produces forecasts of the target variable in future $K$ horizons at a FCT $t$ by fusing previous $W$ observations and future $K$ NWP predictions.

First, we stack a LSTM ($LSTM^{(h)}$) and a MLP ($MLP^{(h)}$) to form a historical encoder to capture historical information as a state vector ($\mathbf{h}^{(h)}_{t}$). Formally:
\begin{equation}
\begin{aligned}
\label{eq:local-history}
\mathbf{h}_{t}   &= LSTM^{(h)}([\mathbf{y}_{[(t-W+1):t]}, \mathbf{x}_{[(t-W+1):t]}]), \\
\mathbf{h}^{(h)}_{t} &= MLP^{(h)}(\mathbf{h}_{t}).
\end{aligned}
\end{equation}
Similar to prior works \citep{wen2017a,wilson2018a}, we found a complicated LSTM tends to overfit time-series data. Thus, $LSTM^{(h)}$ is designed to have one layer and 32 hidden states by default.
$LSTM^{(h)}$ maps the history of sequences to a representation $\mathbf{h}_{t}$ (typically a vector of hidden states at the last step). For each timestamp $t$, we abbreviate the input $[y_{t}, \mathbf{x}_{t}]$ as $\mathbf{x}_{t}$. The formulations of the gates, cell update and output of the LSTM cell are defined as:
\begin{equation}
\begin{aligned}
\label{eq:lstmcell}
\mathbf{i}_{t} &= sigmoid(\mathbf{W}_{ix}\mathbf{x}_{t}+\mathbf{W}_{ih}\mathbf{h}_{t-1} + b_{i}), \\
\mathbf{f}_{t} &= sigmoid(\mathbf{W}_{fx}\mathbf{x}_{t}+\mathbf{W}_{fh}\mathbf{h}_{t-1} + b_{f}), \\
\mathbf{o}_{t} &= sigmoid(\mathbf{W}_{ox}\mathbf{x}_{t}+\mathbf{W}_{oh}\mathbf{h}_{t-1} + b_{o}), \\
\mathbf{c}_{t} &= \mathbf{f}_{t} \odot \mathbf{c}_{t-1} + \mathbf{i}_{t} \odot tanh(\mathbf{W}_{cx}\mathbf{x}_{t}+\mathbf{W}_{ch}\mathbf{h}_{t-1} + b_{c}), \\
\mathbf{h}_{t} &= \mathbf{o}_{t} \odot tanh(\mathbf{c}_{t}), \\
\end{aligned}
\end{equation}
where $\textbf{W}.$ are the weight matrices, $b.$ are the biases, and $\odot$ is the element-wise vector product.
The extracted representation $\mathbf{h}_{t}$ is then fed into $MLP^{(h)}$ to form a higher level representation $\mathbf{h}^{(h)}_{t}$ as output. We implement $MLP^{(h)}$ as a one-layer MLP with a rectified linear unit ($relu$) as the activation function \citep{glorot2011deep}. The number of units is set to be double of the length of the input.
We claim that $MLP^{(h)}$ is necessary for extracting better representations. Intuitively, $\mathbf{h}_{t}$ is inclined to capture the behavior of the input time series, while $\mathbf{h}^{(h)}_{t}$ is apt to reconstruct the input information to align with the future horizons.

Second, the future encoder applies a standard MLP ($MLP^{(f)}$) to encode the NWP predictions in $K$ future horizons as a representation vector $\mathbf{h}^{(f)}_{t+1}$. Formally:
\begin{equation}
\label{eq:local-future}
\mathbf{h}^{(f)}_{t+1} = MLP^{(f)}([\mathbf{y}^{(f)}_{[(t+1):(t+K)]},\mathbf{x}^{(f)}_{[(t+1):(t+K)]}]).
\end{equation}
Here the future encoder turns out to be crucial due to the following reasons: (1) Station observations reflect the local ground truth and can be substantially different from the NWP data; (2) The estimated error in NWP predictions requires special consideration.
Our main motivations to use the MLP are twofold. First, NWP predictions are high level features close to the final output and hence a simple model is enough, while a complex model may mostly twist the features. Second, some prior research works applied MLPs to downscale NWP predictions \citep{krasnopolsky2006complex} or act as function approximators to replace parts of a NWP model for reduced computational complexity \citep{rasp2018deep}. In contrast to extrapolating the historical data, these approaches conduct interpolation or regression on data with well aligned input and output. Therefore, we believe our $MLP^{(f)}$ can produce a representation aligned with the future horizons.
While it is tempting to replace $MLP^{(f)}$ with another LSTM, this replacement brings in negative effects in our case. We hypothesize this inefficiency is due to the following reasons. With the approximated data as input, LSTM is more complicated than MLP and thus tends to overfit spurious variances. However, it is still too simple when compared with the NWP model to capture a complete picture of atmosphere. Meanwhile, the innately estimated error may be enlarged along the long chain of propagation in the LSTM.

Finally, we feed the historical representation ($\mathbf{h}^{(h)}_{t}$) and the future representation ($\mathbf{h}^{(f)}_{t+1}$) into a MLP ($MLP^{(l)}$) to derive local predictions for the current station. Formally:
\begin{equation}
\label{eq:local-combine}
[\hat{y}^{(l)}_{t+1}; ...; \hat{y}^{(l)}_{t+K}] =  MLP^{(l)}([\mathbf{h}^{(h)}_{t}, \mathbf{h}^{(f)}_{t+1}]).
\end{equation}
$MLP^{(l)}$ connects one hidden layer, in which the activation function is $relu$ and the unit size is identical to the input size, to a linear layer outputting $K$ predicted values in a vector.

\subsection{Spatial module}
We propose a spatial module to construct a joint representation over all stations and produce predictions at respective stations. Our main considerations are as follows. 
The case in Fig. \ref{fig:corr-obs}(c) as well as previous results \citep{damousis2004wind,wilson2018a,shi2015convolutional,ezzat2020turbine} reveal the fact that complex dependences always exist in weather observations from closely distributed stations.
These dependencies are dynamic and vary with the change of atmosphere and locations.
Therefore, simply bundling the data from multiple stations as input is unlikely to help, but could only result in a messy dataset that makes the learning process harder or even infeasible.

\begin{figure}
\centering
\includegraphics[width=0.7\textwidth]{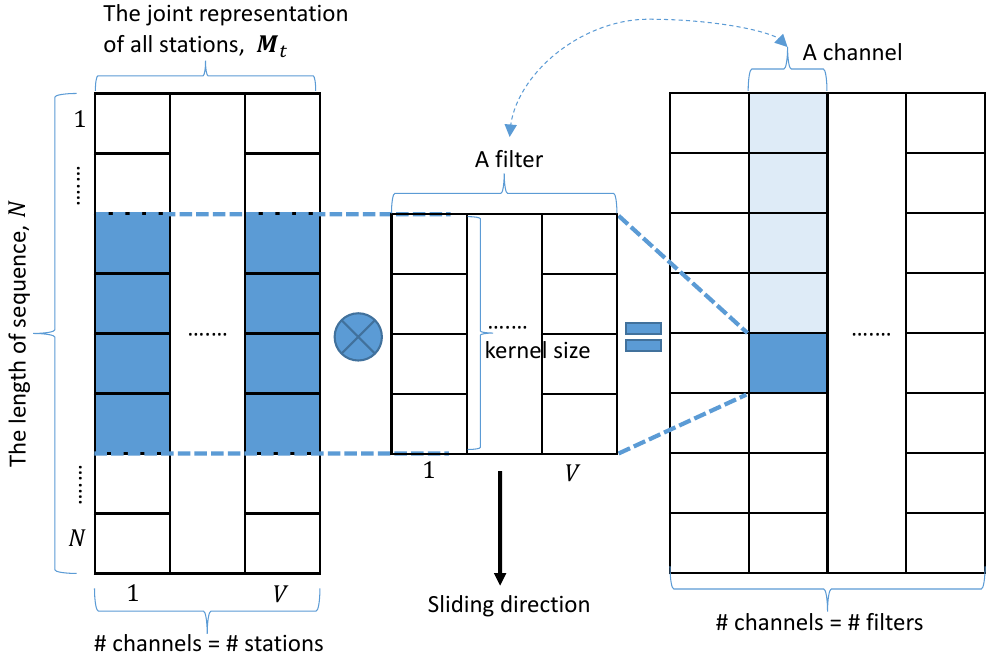}
\caption{The process to distill spatial features in the spatial module, one-dimensional CNN ($CNN^{(s)}$).}
\label{fig:framework-cnn}
\end{figure}

Specifically, for a station $i$, its state at FCT $t$ can be encoded by its temporal module as a representation vector $\mathbf{h}_{t}^{(i)}$, which is exactly the output of the last hidden layer in $MLP^{(l)}$. For readability, the script $i$ to denote a station is omitted without ambiguity. 
We incorporate all the representations in a feature map, represented as $\mathbf{M}_{t} \in \mathbb{R}^{N \times V}$, that has $V$ channels with each corresponds to a station's representation of length $N$. Thus, $\mathbf{M}_{t}$ represents the state of the whole field at FCT $t$. Then, we feed $\mathbf{M}_{t}$ into a one-dimensional CNN ($CNN^{(s)}$) to produce predictions for one station. Formally:
\begin{equation}
\begin{aligned}
\label{eq:local-spatial}
[\hat{y}^{(s)}_{t+1}; ...; \hat{y}^{(s)}_{t+K}] =  CNN^{(s)}(\mathbf{M}_{t}), \\
\mathbf{M}_{t} \in \mathbb{R}^{N \times V} = \{\mathbf{h}_{t}^{(i)}\}_{i=1}^{V}, \mathbf{h}_{t}^{(i)} \in \mathbb{R}^{N}.
\end{aligned}
\end{equation}
Fig.~\ref{fig:framework-cnn} illustrates how $CNN^{(s)}$ captures spatial dependences between stations. Here, a filter is sliding over the feature map $\mathbf{M}_{t}$ to extract a type of features as a channel of values on the output feature map. In the dark blue region, a convolution is operating between the filter and a segment of $\mathbf{M}_{t}$ that spans all stations. In this way, the output feature map resulted from multiple filters can capture various types of spatial features. 
We implement $CNN^{(s)}$ as a stack of a convolutional layer, a max pooling layer, a flatten layer and a linear dense output layer. The convolutional layer takes following settings: the activation function is $relu$, the number of filters is $64$, and the kernel size is $5$. The pool size of the pooling layer is $2$. The dense layer fed with the flatten feature vector makes predictions in future $K$ horizons as a vector.

\subsection{Ensemble module}    
The ensemble module of a station uses a dense layer without the bias parameter to weigh the local predictions ($\hat{\mathbf{y}}^{(l)}_{[(t+1):(t+K)]}$) and the spatial predictions ($\hat{\mathbf{y}}^{(s)}_{[(t+1):(t+K)]}]$) before generating the final predictions ($\hat{\mathbf{y}}_{[(t+1):(t+K)]}$). The dependencies among future horizons are also considered in this module. Formally:
\begin{equation}
\begin{aligned}
\label{eq:combine}
& \hat{\mathbf{y}}_{[(t+1):(t+K)]} = E([\hat{\mathbf{y}}^{(l)}_{[(t+1):(t+K)]}, \hat{\mathbf{y}}^{(s)}_{[(t+1):(t+K)]}]), \\
& \hat{y}_{t+j} = \sum_{i=1}^{K}\mathbf{W}^{(l)}_{ij}\hat{y}^{(l)}_{t+i} + \sum_{i=1}^{K}\mathbf{W}^{(s)}_{ij}\hat{y}^{(s)}_{t+i}, j=1,2, .., K.
\end{aligned}
\end{equation}
We empirically found that the initialization of weights is critical for ensuring prediction accuracy.
Specifically, for a predicted value ($\hat{y}_{t+j}$), we uniformly initialize the weights, $\mathbf{W}^{(l)}_{.j}$ and $\mathbf{W}^{(s)}_{.j}$, to $\frac{1}{2K}$. We believe such a strategy eases the learning process. It could be inferred from Fig.~\ref{fig:corr-obs}(a) that the predicted values are moderately correlated with each other. A uniform initialization, therefore, can produce a set of weights that are closer to the ground truth, at least in comparison with the random or zero initialization.
Besides, we argue that the ensemble module is indispensable to take advantage of both temporal and spatial information. The key reason is that spatial dependences are dynamic along the state of atmosphere \citep{wilson2018a,damousis2004wind} and thus they are not always helpful for a station.

\subsection{Covariate selection module}
\label{sec:app-cov}

To automatically select influential meteorological variables, we devise a covariate selection module that consists of two components to handle the historical and the future data (i.e., the historical observation and the NWP data), respectively. The core idea is to quantify the importance of each covariate and then identify a threshold to pick up the important ones. Different from prior works, we consider the predictive ability of each candidate with regards to the target variable.

For the historical data of a station, $(\mathbf{y} \in \mathbb{R}^{T}, \mathbf{x} \in \mathbb{R}^{T \times D})$, we measure the importances of all covariates as follows.
First, a one-step ahead prediction model, i.e., a ridge regression model \citep{donald1975ridge} without the bias parameter, is trained for each future horizon $k$: $\hat{y}_{t+k} = Ridge^{(k)}(y_{t}, \mathbf{x}_{t}; \mathbf{c}^{(k)})$, $\mathbf{c}^{(k)} \in \mathbb{R}^{D+1}, k=1,...,K$, where $\mathbf{c}^{(k)}$ involves weights of the covariates including the self-history. Second, we consider the magnitude of weights, i.e., let each $\mathbf{c}^{(k)} = \left | \mathbf{c}^{(k)} \right|$.
Third, A min-max normalization is applied to cast each value of $\mathbf{c}^{(k)}$ into the range of $[0,1]$: $\mathbf{c}_{j} = \frac{\mathbf{c}_{j}-min(\mathbf{c})}{max(\mathbf{c})-min(\mathbf{c})}, j=1,...,D+1$.
Finally, for each covariate $j$, an average of its weights over all horizons will be its final importance factor: $\mathbf{c}_{j} = \frac{1}{K}\sum_{k=1}^{K}\mathbf{c}^{(k)}_{j}, j=1,...,D+1$. A respective importance vector, $\mathbf{c}$, can then be produced for the current station.

For the future data of a station, $(\mathbf{y}^{(f)} \in \mathbb{R}^{T}, \mathbf{x}^{(f)} \in \mathbb{R}^{T \times D^{(f)}})$, we apply a similar method to calculate importance factors for the covariates, denoted as $\mathbf{c}^{(f)}$. One exception is that there just a point-to-point ridge regression model is needed: $\hat{y}_{t} = Ridge^{(f)}(y_{t}^{(f)}, \mathbf{x}_{t}^{(f)}; \mathbf{c}^{(f)}), \mathbf{c}^{(f)} \in \mathbb{R}^{D^{(f)}+1}$.

\section{Experiment and Application}
\label{sec:exp}

We firstly introduce the experimental setup, including a real-world dataset, evaluation settings, implementation details and comparison methods. Then, we demonstrate the effectiveness of MHSTN and the contributions of components within the framework by comprehensive experiments. The results encourage an effective unified framework (i.e., MHSTN) to model complicated natures in spatiotemporal weather data.

\subsection{Dataset description}
\begin{figure}
     \centering
     \begin{tabular}[c]{cc}
     \begin{subfigure}[b]{0.35\textwidth}
         \includegraphics[width=\textwidth]{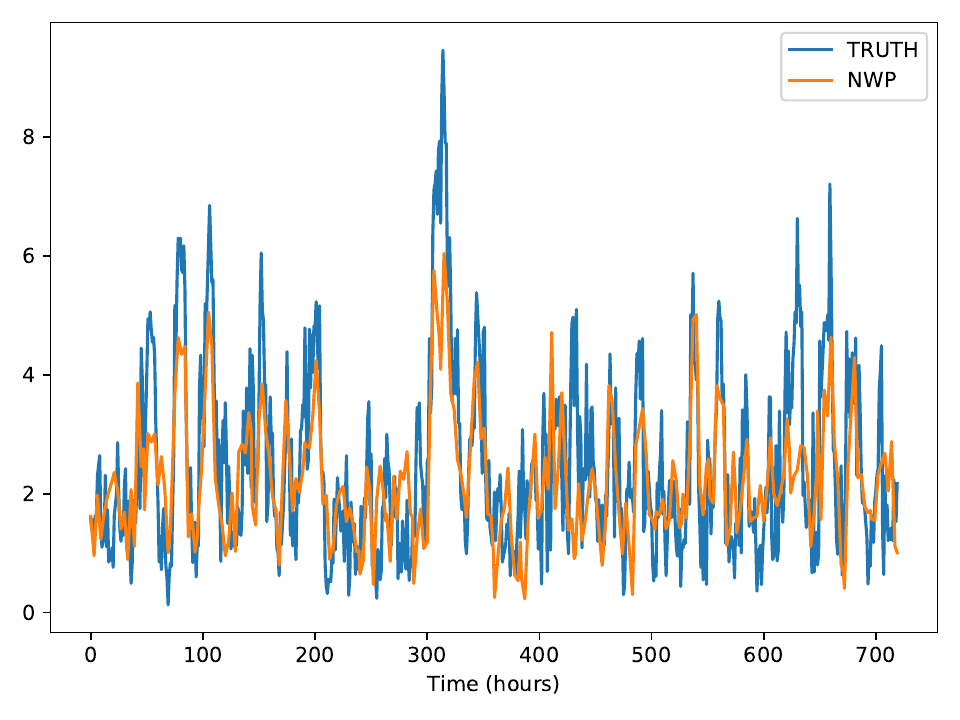}
         \caption{Wind speed ($v$)}
         \label{fig:vis-v}
     \end{subfigure} &
     \begin{subfigure}[b]{0.35\textwidth}
         \includegraphics[width=\textwidth]{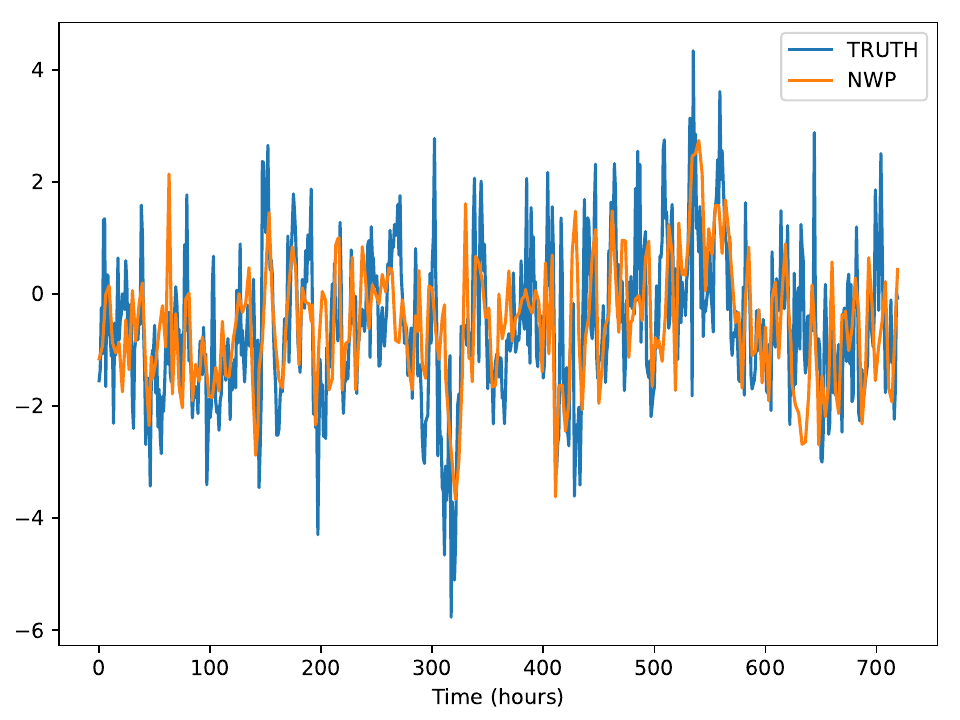}
         \caption{lateral wind speed ($vx$)}
         \label{fig:vis-vx}
     \end{subfigure} \\
     
     \begin{subfigure}[b]{0.35\textwidth}
         \includegraphics[width=\textwidth]{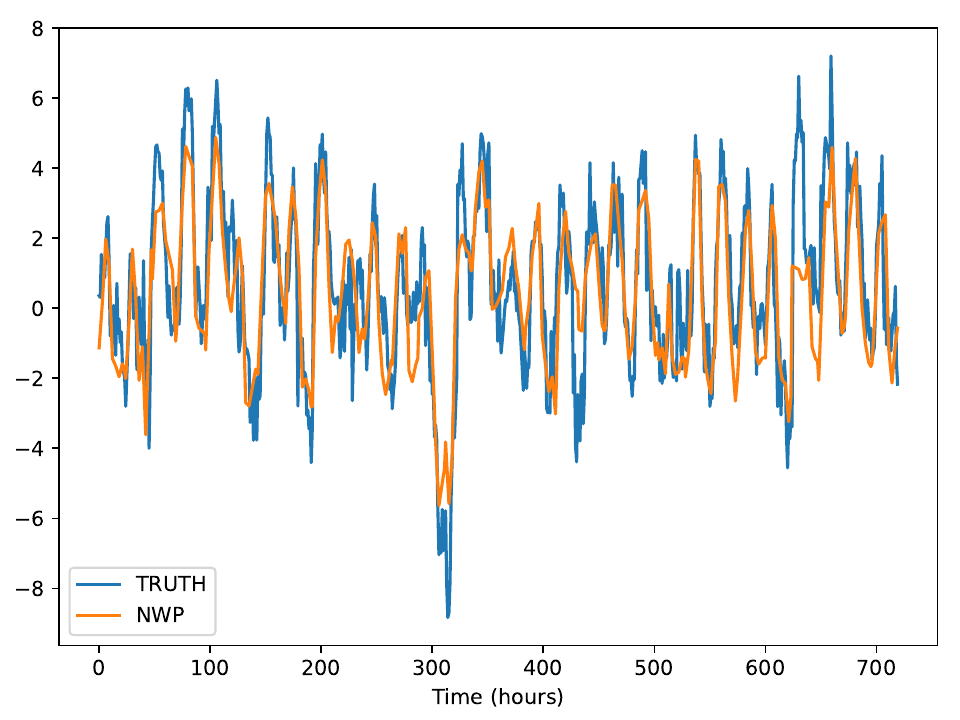}
         \caption{longitudinal wind speed ($vy$)}
         \label{fig:vis-vy}
     \end{subfigure} &
     \begin{subfigure}[b]{0.35\textwidth}
         \includegraphics[width=\textwidth]{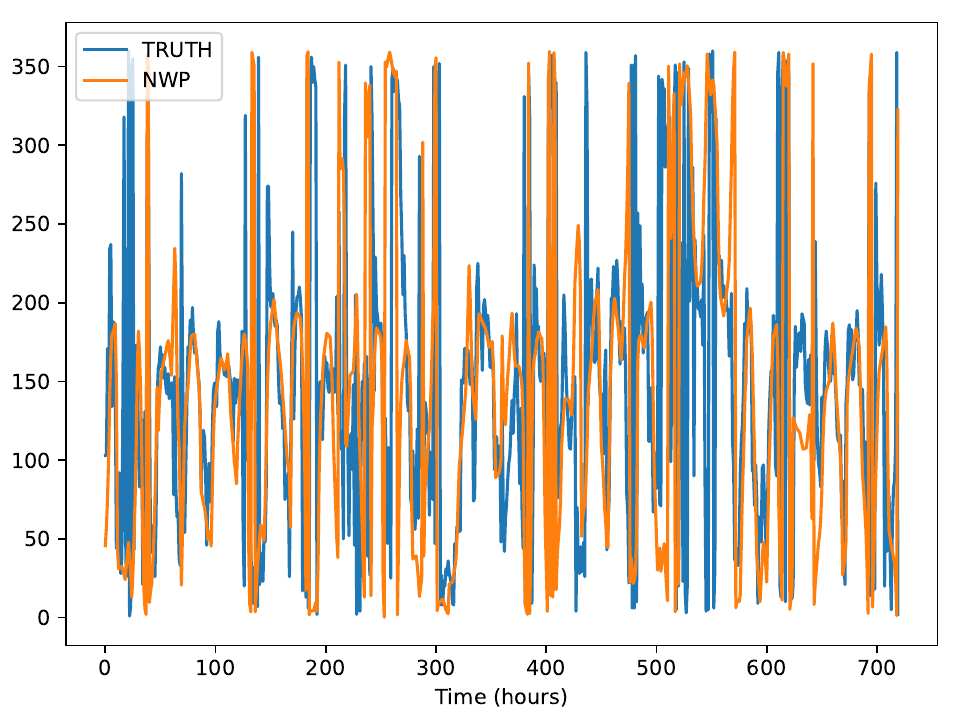}
         \caption{wind direction ($\theta$).}
         \label{fig:vis-dir}
     \end{subfigure} \\
     
     \begin{subfigure}[b]{0.35\textwidth}
         \includegraphics[width=\textwidth]{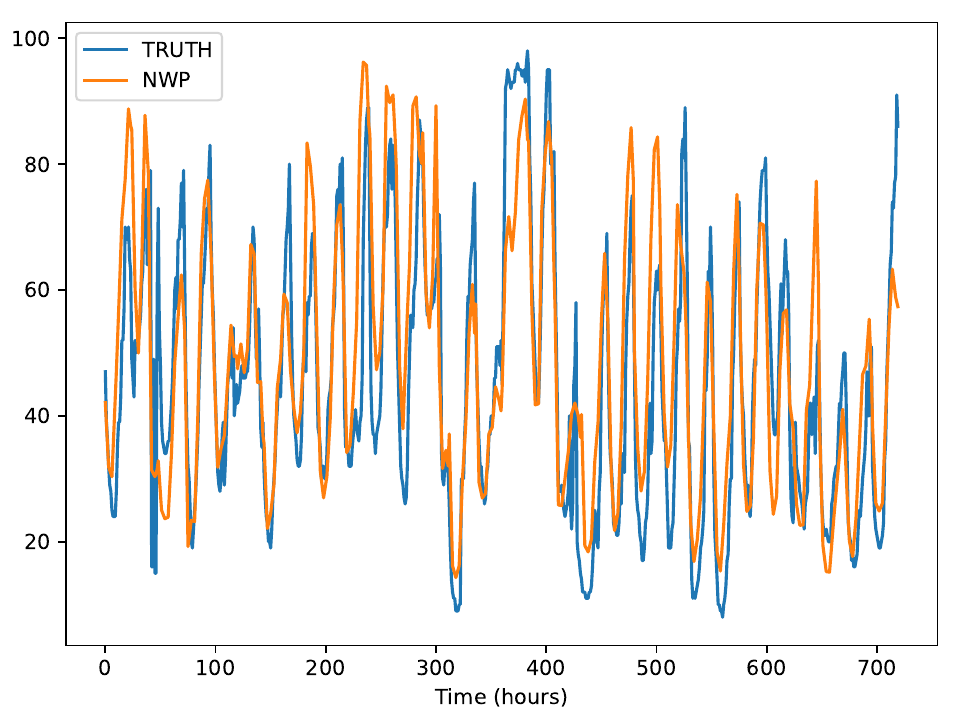}
         \caption{relative humidity ($rh$).}
         \label{fig:vis-rh}
     \end{subfigure} &
     \begin{subfigure}[b]{0.35\textwidth}
         \includegraphics[width=\textwidth]{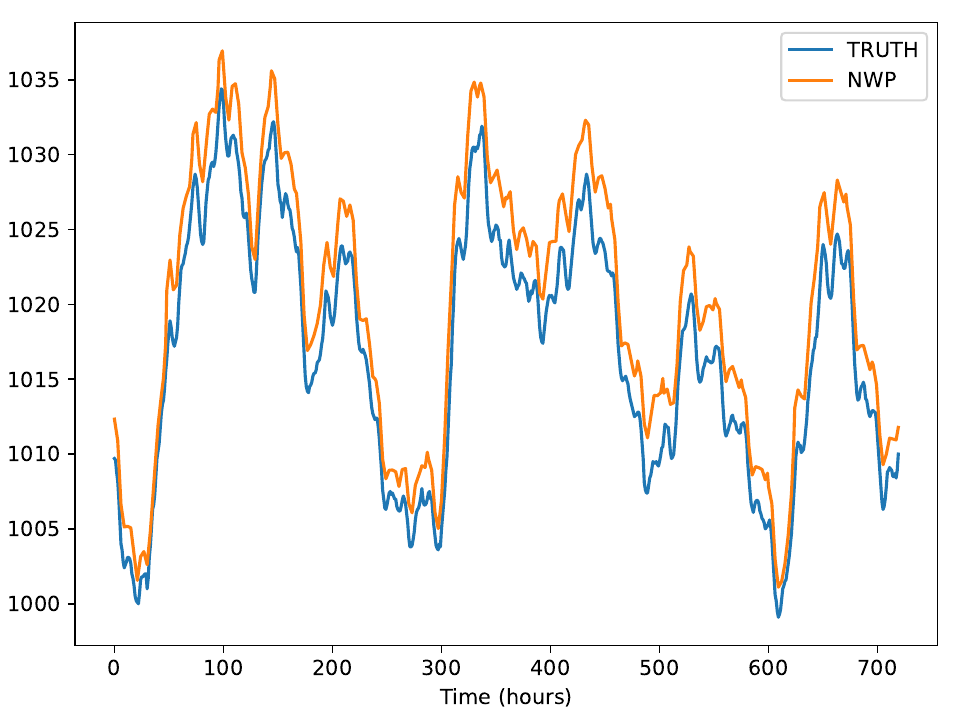}
         \caption{sea-level pressure ($slp$)}
         \label{fig:vis-slp}
     \end{subfigure} \\

     \begin{subfigure}[b]{0.35\textwidth}
        \includegraphics[width=\textwidth]{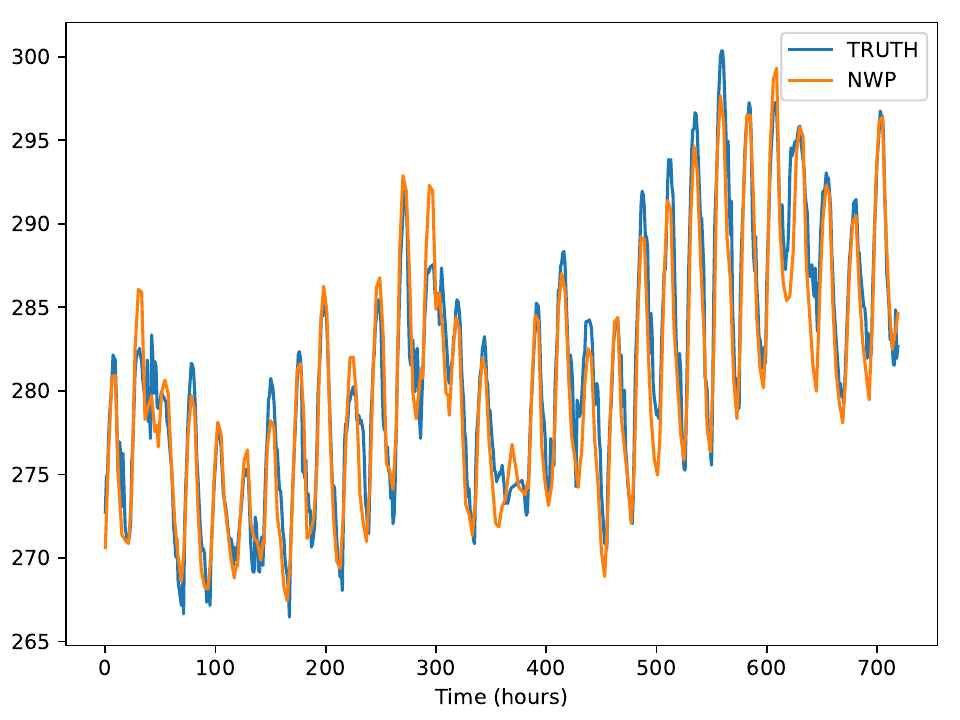}
        \caption{Temperature ($tp$).}
        \label{fig:vis-tp}
    \end{subfigure}
    \end{tabular}    
    \caption{Visualizations of meteorological variables.}
    \label{fig:vis-vars}
\end{figure}

We construct a dataset using real-world data from an international airport (shown in Fig.~\ref{fig:corr-nwp}). The dataset consists of nine groups of observation data collected at nine closely distributed stations and one group of NWP data for the airfield. 
The data are hourly time series of weather variables collected in the period from 02:00:00, Mar. 2018 to 23:00:00, Sep. 2019 with a total of 13,176 time points (24 points per day).
Tab.~\ref{tab:var} illustrates the included meteorological  variables. Note that the values of $vx$ and $vy$ in the observation data are calculated from $v$ and $\theta$ by Eq.~\eqref{eq:decomp-spd} as they cannot be measured directly.
Tab.~\ref{tab:data} illustrates the basic information of the meteorological data. 
Fig.~\ref{fig:vis-vars} visualizes each weather variable over the first 30 days. 
There are missing values in the observation data. For any station, the missing ratio of a target variable (i.e., $v$, $vx$, $vy$ or $\theta$) is less than 0.2\%. We simply follow the common practice of aviation business to fill any missing value with the average of its two closest measurements along the directions of history and future.

\begin{table}[htbp]
\caption{Meteorological variables.}
\label{tab:var}
\centering
\begin{tabular}{cccc}
\toprule
Meteorological variables              & Abb.    &Unit  & Observed altitude (meters)\\  
\midrule
wind speed          & $v$  & m/s  & 10 \\
lateral wind speed  & $vx$ & m/s  & -\\
longitudinal wind speed & $vy$ & m/s & - \\
wind direction & $\theta$ & degree & 10 \\
temperature & $tp$ & K  & 2\\
relative humidity & $rh$ & \%  & 2\\
sea-level pressure & $slp$ & hpa & 2\\
\bottomrule
\end{tabular}
\end{table}
\begin{table}[htbp]
\caption{Basic information of meteorological data.}
\label{tab:data}
\centering
\begin{tabular}{l|cccc|cccc}
\toprule
 & \multicolumn{4}{c}{Observation data}  & \multicolumn{4}{c}{NWP data} \\
Abb. &Min  &Mean  &Max &Std  &Min  &Mean  &Max &Std\\  
\midrule
$v$  &0.15   &2.93 &14.97 &1.80 & 0.04 & 2.52 & 10.96 & 1.37   \\
$vx$ &-10.40 & -0.08 & 12.18 & 2.09 & -6.61 & 0.090 & 9.89 & 1.65\\
$vy$ &-12.53 &-0.26 &9.16 &2.72 &-8.34 &-0.14 &7.56 &2.25  \\
$\theta$ &1.00 & 177.66 &360.00 &112.98 &0.07 &176.89 &359.93 &109.24  \\
$tp$ &259.95 &289.43 &312.99 &11.73 &259.92 &288.32 &312.52 &11.75 \\
$rh$ &5.44 &49.81 &99.11 &24.45 &4.09 &50.39 &99.88 &25.12 \\
$slp$ &990.54 &1010.91 &1043.61 &10.55 &993.28 &1014.55 &1046.08 &10.80 \\
\bottomrule
\end{tabular}
\end{table}

\subsection{Evaluation settings}
To extensively evaluate the robustness of prediction models, we follow the rolling and incremental evaluation strategies in time series prediction \citep{guo2018bitcoin}. Such evaluation strategies are similar to but different from the standard cross-validation methods that the successiveness of time series data is maintained. 
In particular, we divide the whole time range of data into non-overlapping intervals with each corresponds to one month. 
In the \textbf{rolling evaluation}, we take data within one interval as the testing set and the data in the previous $I$ intervals as the training set where the last interval is held out as the validation set if needed. For comparison, the \textbf{incremental evaluation} uses all the data preceding to the current testing set for training. 
In our application, there are $18$ months/intervals (a year and a half) of data. We set the parameter $I$ in the rolling evaluation to $6$. For either evaluation, we slid a window corresponding to the current testing set over the last 12 intervals and thus conduct $12$-fold validations that span across all months/seasons in a year.
In a fold of validations, we follow the common practice in machine learning domain to report average performance over multiple runs with different random seeds (i.e., $10$ random runs) because it is recognized that a good neural network should usually be robust to different initializations. 

We use Root Mean Square Error (RMSE), i.e., the sqrt of MSE defined in Eq.~\eqref{eq:loss}, as the metrics to report the prediction performance on the wind speed variables, i.e., $v$, $vx$ and $vy$. Basically, there is no substantial difference between MSE and RMSE except that the former is convenient for derivation while the latter is commonly used as an evaluation metrics. 
Our motivation to choose MSE/RMSE is that they are more sensitive to extreme values than other losses/metrics do. Extreme values, which are frequently treated as noise in other application domains, reflect extreme weather events that are usually of importance. In our application to an international airport, RMSE is the critical metrics to evaluate candidate algorithms for wind speed prediction because the greater the wind speed implies the greater the risk.
We apply an Adjusted Mean Absolute Error (AMAE) \citep{grimit2006the}, defined as Eq.~\eqref{eq:mae-dir}, for the circular variable of $\theta$ due to the RMSE is no longer feasible. Suppose a case involving a wind direction of $10$ degree. There are two candidate predicted values, i.e., $350$ and $40$. RMSE and common MAE will pick up $40$, while $350$ is closer to the ground truth, $10$, in a circular coordinate system. 
\begin{equation}
\begin{aligned}
\label{eq:mae-dir}
AMAE &= \frac{1}{T}\sum_{t=1}^{T}\delta(\hat{\theta}_{t}, \theta_{t}), \\
\delta(\hat{\theta}_{t}, \theta_{t}) &= \left\{\begin{matrix}
|\hat{\theta}_{t}-\theta_{t}| & if\ 0 < |\hat{\theta}_{t}-\theta_{t}| \leqslant 180, \\
(360-|\hat{\theta}_{t}-\theta_{t}|) & if\ 180 < |\hat{\theta}_{t}-\theta_{t}| \leqslant 360.
\end{matrix}\right.
\end{aligned}
\end{equation}

We also report relative/normalized errors for easy comparison. For wind speed variables (i.e., $v$, $vx$ and $vy$), we report Root Relative Squared Error (RRSE). RRSE is the root of the squared error of the predictions relative to a naive model that always outputs the average of the actual values. Eq.~\eqref{eq:rrse} presents the mathematical definition, where $y_{t}$ denotes observation value, $\hat{y_{t}}$ denotes prediction value and $\bar{y}$ is the average of all $T$ observations. 
For $\theta$, we report Normalized AMAE (NAMAE) that is equal to divide AMAE by the maximum error $180$ degree (i.e., $NAMAE = AMAE/180$). 
It should be clarified that NAMAE presents a percentage error ranging in [0, 1]. Although RRSE is not limited to an upper bound of 1, it is relative to a naive model and thus is also comparable across different settings. In addition, the square root of the RSE reduces the error to the same dimension as the quantity being predicted. 
\begin{equation}
\label{eq:rrse}
RRSE = \sqrt{\frac{\sum_{t=1}^{T}(y_{t}-\hat{y}_{t})^{2}}{\sum_{t=1}^{T}(y_{t}-\bar{y})^{2}}}
\end{equation}

\subsection{Implementation details}

To meet the practical requirements, the number of future horizons $K$ is set to 24 (hours) and the FCT $t$ is 00 (o'clock). Due to the latest auto-correlations are far more significant, as exemplified in Fig.~\ref{fig:corr-obs}(a), we set the historical window $W$ to the minimum period, i.e., 24 (hours).

We train a MHSTN for each of wind speed variables (i.e., $v, vx$ and $vy$) over all stations with following three stages. We first train the temporal modules for all stations, respectively. Then we train the spatial module of each station. Third, the ensemble modules of all stations are trained. In the end, the predictions of wind direction ($\theta$) can be efficiently produced by Eq.~\eqref{eq:spd2degree}.

We implement DNNs with TensorFlow-Keras-2.3.0 and train them by Adam algorithm \citep{kingma2014adam} with the default settings. The initial learning rate ($0.001$) will be reduced by a factor of $0.5$, if no improvement is seen for $3$ epochs on the validation set, and $0.0001$ is the lower bound. The batch size is $32$. We perform early-stopping on the validation set to prevent overfitting, if the loss is not improved for $30$ epochs, and save the best model for testing. 
During training, we normalize the data before feeding them into the model. Specifically, each time series is subtracted by its mean value and then divided by the standard deviation of the training set. After training, we perform an inverse-normalization on the model output to produce final predictions, and report evaluation metrics in the original data space. 
We run experiments on a Ubuntu-18.04 server featuring a 2.50GHz Intel(R) Xeon(R) E5-2682 CPU and a NVIDIA Tesla P100 GPU.

\subsection{Comparison methods}

Our unified model has the capability to assimilate multi-source data related to \textbf{history (h)}, \textbf{future (f)}, \textbf{space (s)} and \textbf{covariates (c)} (excepting self-history and self-future). For comparison, the models fed with all available covariates are marked with \textbf{c*}.
The comparison methods include advanced physical model, conventional machine learning models, deep neural networks, and variants of our model. 

First, we consider two common benchmarks in weather prediction as follows: 
\begin{itemize}
    \item \textbf{Persistence} model predicts the wind results by simply copying the results on 24 hours ago. It has been popularly applied to characterize the difficulty of forecasting \citep{wilson2018a}. 
    \item \textbf{NWP} is the commercial NWP service in our application. It should be noted that NWP is usually a hard-to-beat physical model \citep{bauer2015the}.
\end{itemize}

In addition, we implement two conventional regression models, which have been widely used in short-term forecasting problems, as follows:
\begin{itemize} 
    \item \textbf{SVR} indicates the Support Vector Regression model which is an extension from the non-linear support vector machine for regression estimation. 
    \item \textbf{GBRT} indicates the Gradient Boosting Regression Tree which is an ensemble model for regression tasks and have been widely exploited in practice.
\end{itemize}
Specifically, our implementation is based on scikit-learn-0.23.1. The above regression models just can output a value once a time, thus we warp them into the module of MultiOutputRegressor \footnote{\url{https://scikit-learn.org/stable/modules/generated/sklearn.multioutput.MultiOutputRegressor.html}} that is an implementation of the direct strategy for multi-horizon forecasting (please refer to section~\ref{sec:rel} for more details). We also extensively search crucial hyper-parameters with randomized search \footnote{\url{https://scikit-learn.org/stable/modules/generated/sklearn.model_selection.RandomizedSearchCV.html}} and save the best models for comparison.

In the following content, we include a series of DNN models.

First, \textbf{MLP} and \textbf{LSTM} networks have been mostly used for wind speed prediction. Especially, we adapt components in MHSTN's temporal module as competitors because we found a more complicated network cannot bring in any benefits in terms of prediction accuracy. 
\begin{itemize}
    \item \textbf{LSTM(h)}, \textbf{LSTM(f)} and \textbf{LSTM(h,f)}: Each of them is a LSTM with different inputs that connects a network of $LSTM^{(h)}$ in Fig.~\ref{fig:framework} to a linear output layer. 
    \item \textbf{MLP(h)}, \textbf{MLP(f)}, and \textbf{MLP(h,f)}: Each of them is a MLP with different inputs that  concatenates a network of $MLP^{(f)}$ in Fig.~\ref{fig:framework} to an output layer.
\end{itemize}
Second, we consider the neural networks that has the capability to model spatiotemporal data.
\begin{itemize}
    \item \textbf{ConvLSTM(h,f,s)}: Convolutional LSTM was originally proposed by \cite{shi2015convolutional} to model spatiotemporal sequences for precipitation nowcasting. The network incorporates convolutional structures to model both input-to-state and state-to-state transitions.
    We apply a standard ConvLSTM with an input of 4-D tensor, i.e., (\textit{samples}, \textit{time}, \textit{rows}, \textit{channels}), in which \textit{rows} is $2$ corresponding to the two sequences of history and future at a station and \textit{channels} is the number of stations. We implement the model by keras \footnote{\url{https://www.tensorflow.org/api_docs/python/tf/keras/layers/ConvLSTM2D}} with settings according to our spatial module (i.e., $CNN^{(s)}$)). 
    \item \textbf{GCNLSTM(h,f,s)}: A coupled Graph Convolutional Network (GCN) \citep{lipf2017semisupervised} and LSTM has been applied to model spatiotemporal data for one-step ahead wind speed prediction \citep{khodayar2019spatiotemporal,wilson2018a} and traffic prediction \citep{zhao2019tgcn}. 
    Similarly, we implement a GCNLSTM that has three blocks as follows: (1) A GCN block, which is similar to our spatial module (i.e., $CNN^{(s)}$) and relies on the spektral library \footnote{\url{https://github.com/danielegrattarola/spektral}}; (2) a LSTM block, which is the same as our historical encoder (i.e., $LSTM^{(h)}$); and (3) a MLP block, which has a hidden layer with units identical to the length of input and a linear output layer to make predictions. The outputs of the first two blocks are concatenated and fed into the third block. All hidden layers in GCNLSTM adopt a \textit{relu} activation function.
    GCNLSTM takes two inputs: a multivariate time series that is a 2-D matrix involving historical and future sequences over all stations (fed into both GCN and LSTM blocks), and an adjacency matrix describing the dynamic spatial dependencies between stations (fed into GCN only), which is a Pearson correlation matrix calculated on the previous input.
\end{itemize}

We also include the Deep Uncertainty Quantification neural network (\textbf{DUQ}) lately proposed by \cite{wang2019deep}, which combines historical observation and NWP data for wind speed prediction at stations deployed in a coarse grid. With different inputs, there are \textbf{DUQ(h,f)}, \textbf{DUQ(h,f,c)} and \textbf{DUQ(h,f,c*)}. The last one corresponds to the proposed DUQ in \citep{wang2019deep}. Specifically, we implement DUQ with the opened source code \footnote{\url{https://github.com/BruceBinBoxing/Deep_Learning_Weather_Forecasting}}, in which we use the proposed negative log-likelihood error (NLE) loss function and one layer Seq2Seq with $32$ hidden notes. There are several notifications should be clarified as follows. (1) We found different settings of DUQ have no significant differences in terms of prediction accuracy and thus picked the current setting. (2) We did not consider exhaustive ensemble strategies due to it is a generally practice to boost the prediction accuracy. (3) DUQ did not leverage spatial dependences and select covariates. 

Finally, we consider variants of our MHSTN (Fig.~\ref{fig:framework}). There are the temporal module, \textbf{MHSTN-T(h,f)}, the spatial module, \textbf{MHSTN-S(h,f,s)}, the ensemble module, \textbf{MHSTN-E(h,f,s)}, and the whole framework, \textbf{MHSTN-E(h,f,s,c)}. 
Furthermore, we also feed covariates into the aforementioned competitors for comparison. 
Interested readers please refer to our source code for more details. 

\subsection{Experimental results}

In this section, we summarize major experimental results in Tab.~\ref{tab:res-roll}, Tab.~\ref{tab:res-roll-norm}, Tab.~\ref{tab:res-inc} and Tab.~\ref{tab:res-inc-norm}. Each reported value is the average error over 9 (stations) $\times$ 12 (intervals/months) $\times$ 10 (random runs) tests. Our framework and it's modules as well as the best results are marked in gray front. 
Furthermore, we analyze covariates' importances, visualize some cases, and report MHSTN's computation time.

\begin{table}[htbp]
\caption{Results of the rolling evaluation}
\label{tab:res-roll}
\centering
\renewcommand\tabcolsep{5.0pt}
\begin{tabular}{p{2cm}p{1.75cm}p{1.75cm}p{1.75cm}p{1.9cm}}
\toprule
              & \multicolumn{4}{c}{Target wind variables}  \\
              \cline{2-5}
Models              & $v$ (RMSE)    &$vx$ (RMSE)   & $vy$ (RMSE)    & $\theta$ (AMAE) \\  
\midrule
\multicolumn{5}{c}{Benchmarks}     \\
\hline
Persistence    &2.322  &2.720  &3.256   &68.889  \\
NWP            &1.516  &1.577  &1.701   &40.124   \\
\midrule
\midrule
\multicolumn{5}{c}{Ingestion of historical(h) and/or future(f) sequences}     \\
\hline
LSTM(h)        &1.800 $\pm$ 0.031   &2.068 $\pm$ 0.031   &2.526 $\pm$ 0.031   &66.571 $\pm$ 1.505 \\
LSTM(f)        &1.632 $\pm$ 0.079   &1.665 $\pm$ 0.079   &1.913 $\pm$ 0.079   &44.299 $\pm$ 1.225 \\
LSTM(h,f)      &1.618 $\pm$ 0.068   &1.706 $\pm$ 0.068   &1.953 $\pm$ 0.068   &44.420 $\pm$ 1.459 \\
MLP(h)         &1.852 $\pm$ 0.035   &2.105 $\pm$ 0.035   &2.639 $\pm$ 0.035   &69.253 $\pm$ 1.757 \\
MLP(f)         &1.399 $\pm$ 0.019   &1.536 $\pm$ 0.019   &1.731 $\pm$ 0.019   &41.541 $\pm$ 0.800 \\
MLP(h,f)       &1.445 $\pm$ 0.029   &1.628 $\pm$ 0.029   &1.831 $\pm$ 0.029   &43.989 $\pm$ 1.156 \\
\hline
\multicolumn{1}{>{\columncolor{mygray}}l}{MHSTN-T(h,f)} 
               &1.401 $\pm$ 0.024   &1.533 $\pm$ 0.024 &1.687 $\pm$ 0.024   &40.506 $\pm$ 0.805 \\
\hline
GBRT(f)         &1.444 $\pm$ 0.027  &1.608 $\pm$ 0.027 &1.774 $\pm$ 0.027  &41.608 $\pm$ 0.791  \\
GBRT(h,f)       &1.440 $\pm$ 0.027  &1.598 $\pm$ 0.027 &1.742 $\pm$ 0.027  &40.648 $\pm$ 0.757 \\
SVR(f)          &1.447 $\pm$ 0.010  &1.622 $\pm$ 0.010 &1.771 $\pm$ 0.010  &40.946 $\pm$ 0.290  \\
SVR(h,f)        &1.456 $\pm$ 0.010  &1.679 $\pm$ 0.010 &1.847 $\pm$ 0.010  &42.719 $\pm$ 0.313 \\
DUQ(h,f)        &1.696 $\pm$ 0.006  &2.055 $\pm$ 0.006 &2.628 $\pm$ 0.006  &79.310 $\pm$ 0.746 \\
\midrule
\midrule
\multicolumn{5}{c}{Addition of spatial information (s)}     \\
\hline
\multicolumn{1}{>{\columncolor{mygray}}l}{MHSTN-S(h,f,s)} 
               &1.381 $\pm$ 0.019 
               &\multicolumn{1}{>{\columncolor{mygray}}l}{1.503 $\pm$ 0.019} 
               &1.672 $\pm$ 0.019  
               &39.707 $\pm$ 0.667\\
\multicolumn{1}{>{\columncolor{mygray}}l}{MHSTN-E(h,f,s)}
               &1.370 $\pm$ 0.014 
               &1.504 $\pm$ 0.014  
               &\multicolumn{1}{>{\columncolor{mygray}}l}{1.658 $\pm$ 0.014} 
               &\multicolumn{1}{>{\columncolor{mygray}}l}{39.441 $\pm$ 0.489}\\
\hline
ConvLSTM(h,f,s) &1.378 $\pm$ 0.017 &1.528 $\pm$ 0.017 &1.699 $\pm$ 0.017  &39.885 $\pm$ 0.617\\
GCNLSTM(h,f,s)  &1.464 $\pm$ 0.047 &1.603 $\pm$ 0.047 &1.788 $\pm$ 0.047  &42.410 $\pm$ 1.311\\
\midrule
\midrule
\multicolumn{5}{c}{Addition of covariates (c)}     \\
\hline
\multicolumn{1}{>{\columncolor{mygray}}l}{MHSTN-E(h,f,s,c)}  
     & \multicolumn{1}{>{\columncolor{mygray}}l}{1.365 $\pm$ 0.017}  
     &1.518 $\pm$ 0.017 
     &1.667 $\pm$ 0.017   
     &39.886 $\pm$ 0.486 \\
\multicolumn{1}{l}{MHSTN-E(h,f,s,c*)}    &1.553 $\pm$ 0.034  &1.761 $\pm$ 0.034  &1.924 $\pm$ 0.034   &49.009 $\pm$ 0.984 \\ 
\hline
ConvLSTM(h,f,s,c) &1.612 $\pm$ 0.061  &1.565 $\pm$ 0.061 &1.722 $\pm$ 0.061  &41.098 $\pm$ 0.715\\
ConvLSTM(h,f,s,c*) &1.627 $\pm$ 0.053 &1.815 $\pm$ 0.053 &2.038 $\pm$  0.053  &49.866 $\pm$ 1.991 \\
DUQ(h,f,c)        &1.767 $\pm$ 0.045  &2.063 $\pm$ 0.045 &2.633 $\pm$ 0.045 &80.006 $\pm$ 0.922  \\
DUQ(h,f,c*)        &1.764 $\pm$ 0.019  &2.121 $\pm$ 0.019 &2.703 $\pm$ 0.019 &86.745 $\pm$ 2.621  \\
\bottomrule
\end{tabular}
\end{table}

\begin{table}[htbp]
\caption{Results of the rolling evaluation with normalized metrics}
\label{tab:res-roll-norm}
\centering
\renewcommand\tabcolsep{5.0pt}
\begin{tabular}{p{2cm}p{1.75cm}p{1.75cm}p{1.75cm}p{1.9cm}}
\toprule
              & \multicolumn{4}{c}{Target wind variables}  \\
              \cline{2-5}
Models              & $v$ (RRSE)    &$vx$ (RRSE)   & $vy$ (RRSE)    & $\theta$ (NAMAE) \\  
\midrule
\multicolumn{5}{c}{Benchmarks}     \\
\hline
Persistence    &1.280  &1.316  &1.285   &0.383  \\
NWP            &0.863  &0.781  &0.688   &0.223   \\
\midrule
\midrule
\multicolumn{5}{c}{Ingestion of historical(h) and/or future(f) sequences}     \\
\hline
LSTM(h)        &0.995 $\pm$ 0.017   &1.000 $\pm$ 0.017   &0.998 $\pm$ 0.017   &0.370 $\pm$ 0.008 \\
LSTM(f)        &0.910 $\pm$ 0.042   &0.810 $\pm$ 0.042   &0.763 $\pm$ 0.042   &0.246 $\pm$ 0.007 \\
LSTM(h,f)      &0.902 $\pm$ 0.037   &0.829 $\pm$ 0.037   &0.778 $\pm$ 0.037   &0.247 $\pm$ 0.008 \\
MLP(h)         &1.027 $\pm$ 0.019   &1.021 $\pm$ 0.019   &1.039 $\pm$ 0.019   &0.385 $\pm$ 0.010 \\
MLP(f)         &0.795 $\pm$ 0.011   &0.755 $\pm$ 0.011   &0.691 $\pm$ 0.011   &0.231 $\pm$ 0.004 \\
MLP(h,f)       &0.817 $\pm$ 0.016   &0.799 $\pm$ 0.016   &0.729 $\pm$ 0.016   &0.244 $\pm$ 0.006 \\
\hline
\multicolumn{1}{>{\columncolor{mygray}}l}{MHSTN-T(h,f)} 
               &0.793 $\pm$ 0.014   &0.754 $\pm$ 0.014 &0.674 $\pm$ 0.014   &0.225 $\pm$ 0.004 \\
\hline
GBRT(f)         &0.813 $\pm$ 0.014  &0.786 $\pm$ 0.014 &0.710 $\pm$ 0.014  &0.231 $\pm$ 0.004  \\
GBRT(h,f)       &0.810 $\pm$ 0.015  &0.781 $\pm$ 0.015 &0.697 $\pm$ 0.015  &0.226 $\pm$ 0.004 \\
SVR(f)          &0.808 $\pm$ 0.006  &0.789 $\pm$ 0.006 &0.707 $\pm$ 0.006  &0.227 $\pm$ 0.002  \\
SVR(h,f)        &0.812 $\pm$ 0.005  &0.815 $\pm$ 0.005 &0.734 $\pm$ 0.005  &0.237 $\pm$ 0.002 \\
DUQ(h,f)        &0.945 $\pm$ 0.003  &0.994 $\pm$ 0.003 &1.050 $\pm$ 0.003  &0.441 $\pm$ 0.004 \\
\midrule
\midrule
\multicolumn{5}{c}{Addition of spatial information (s)}     \\
\hline
\multicolumn{1}{>{\columncolor{mygray}}l}{MHSTN-S(h,f,s)} 
               &0.783 $\pm$ 0.011
               &0.739 $\pm$ 0.011 
               &0.668 $\pm$ 0.011  
               &0.221 $\pm$ 0.004\\
\multicolumn{1}{>{\columncolor{mygray}}l}{MHSTN-E(h,f,s)}
               &0.776 $\pm$ 0.008
               &\multicolumn{1}{>{\columncolor{mygray}}l}{0.739 $\pm$ 0.008} 
               &\multicolumn{1}{>{\columncolor{mygray}}l}{0.663 $\pm$ 0.008} 
               &\multicolumn{1}{>{\columncolor{mygray}}l}{0.219 $\pm$ 0.003}\\
\hline
ConvLSTM(h,f,s) &0.782 $\pm$ 0.009 &0.749 $\pm$ 0.009 &0.678 $\pm$ 0.009  &0.222 $\pm$ 0.003\\
GCNLSTM(h,f,s)  &0.826 $\pm$ 0.026 &0.786 $\pm$ 0.026 &0.714 $\pm$ 0.026  &0.236 $\pm$ 0.007\\
\midrule
\midrule
\multicolumn{5}{c}{Addition of covariates (c)}     \\
\hline
\multicolumn{1}{>{\columncolor{mygray}}l}{MHSTN-E(h,f,s,c)}  
     & \multicolumn{1}{>{\columncolor{mygray}}l}{0.769 $\pm$ 0.009}  
     &0.745 $\pm$ 0.009 
     &0.666 $\pm$ 0.009   
     &0.222 $\pm$ 0.003 \\
\multicolumn{1}{l}{MHSTN-E(h,f,s,c*)}    &0.867 $\pm$ 0.019  &0.864 $\pm$ 0.019  &0.772 $\pm$ 0.019   &0.272 $\pm$ 0.005 \\ 
\hline
ConvLSTM(h,f,s,c) &0.903 $\pm$ 0.034  &0.766 $\pm$ 0.034 &0.688 $\pm$ 0.034  &0.228 $\pm$ 0.004\\
ConvLSTM(h,f,s,c*) &0.911 $\pm$ 0.030 &0.887 $\pm$ 0.030 &0.817 $\pm$  0.030  &0.277 $\pm$ 0.011 \\
DUQ(h,f,c)        &0.982 $\pm$ 0.024 &0.997 $\pm$ 0.024 &1.052 $\pm$ 0.024 &0.444 $\pm$ 0.005 \\
DUQ(h,f,c*)        &0.982 $\pm$ 0.011  &1.026 $\pm$ 0.011 &1.079 $\pm$ 0.011 &0.482 $\pm$ 0.015  \\
\bottomrule
\end{tabular}
\end{table}

\begin{table}[htbp]
\caption{Results of the incremental evaluation}
\label{tab:res-inc}
\centering
\renewcommand\tabcolsep{5.0pt}
\begin{tabular}{p{2.6cm}p{1.75cm}p{1.75cm}p{1.75cm}p{1.9cm}}
\toprule
              & \multicolumn{4}{c}{Target wind variables}  \\
              \cline{2-5}
Models              & $v$ (RMSE)    &$vx$ (RMSE)   & $vy$ (RMSE)    & $\theta$ (AMAE) \\  
\midrule
\multicolumn{5}{c}{Benchmarks}     \\
\hline
Persistence    &2.322  &2.720  &3.256   &68.889  \\
NWP            &1.516  &1.577  &1.701   &40.124   \\
\midrule
\midrule
\multicolumn{5}{c}{Ingestion of historical(h) and/or future(f) sequences}     \\
\hline
LSTM(h)        &1.753 $\pm$ 0.032   &2.040 $\pm$ 0.016   &2.475 $\pm$ 0.027   &61.920 $\pm$ 1.121 \\
LSTM(f)        &1.530 $\pm$ 0.092   &1.587 $\pm$ 0.037   &1.770 $\pm$ 0.048   &40.726 $\pm$ 0.945 \\
LSTM(h,f)      &1.516 $\pm$ 0.067   &1.615 $\pm$ 0.038   &1.807 $\pm$ 0.052   &41.415 $\pm$ 1.144 \\
MLP(h)         &1.773 $\pm$ 0.032   &2.042 $\pm$ 0.035   &2.508 $\pm$ 0.038   &64.694 $\pm$ 1.500 \\
MLP(f)         &1.362 $\pm$ 0.019   &1.494 $\pm$ 0.020   &1.643 $\pm$ 0.019   &39.403 $\pm$ 0.629 \\
MLP(h,f)       &1.386 $\pm$ 0.025   &1.561 $\pm$ 0.030   &1.705 $\pm$ 0.031   &41.180 $\pm$ 0.953 \\
\hline
\multicolumn{1}{>{\columncolor{mygray}}l}{MHSTN-T(h,f)} 
              &1.346 $\pm$ 0.020   &1.485 $\pm$ 0.027 &1.611 $\pm$ 0.023   &38.497 $\pm$ 0.734\\
\hline
GBRT(f)         &1.391 $\pm$ 0.023  &1.557 $\pm$ 0.031 &1.705 $\pm$ 0.027 &39.544 $\pm$ 0.684  \\
GBRT(h,f)        &1.387 $\pm$ 0.025  &1.541 $\pm$ 0.030 &1.672 $\pm$ 0.025 &38.730 $\pm$ 0.633  \\
SVR(f)         &1.411 $\pm$ 0.010  &1.581 $\pm$ 0.008 &1.680 $\pm$ 0.008 &38.960 $\pm$ 0.238  \\
SVR(h,f)        &1.422 $\pm$ 0.008 &1.640 $\pm$ 0.009 &1.739 $\pm$  0.008  &40.269 $\pm$ 0.282 \\
DUQ(h,f)        &1.670 $\pm$ 0.007 &1.986 $\pm$ 0.005 &2.632 $\pm$ 0.005   &66.067 $\pm$  0.936\\
\midrule
\midrule
\multicolumn{5}{c}{Addition of spatial information (s)}     \\
\hline
\multicolumn{1}{>{\columncolor{mygray}}l}{MHSTN-S(h,f,s)}
               &1.332 $\pm$ 0.016 &1.466 $\pm$ 0.015 &1.601 $\pm$ 0.021  &37.966 $\pm$ 0.542\\
\multicolumn{1}{>{\columncolor{mygray}}l}{MHSTN-E(h,f,s)}&1.319 $\pm$ 0.012 &\multicolumn{1}{>{\columncolor{mygray}}l}{1.466 $\pm$ 0.013}  &\multicolumn{1}{>{\columncolor{mygray}}l}{1.589 $\pm$ 0.011} & \multicolumn{1}{>{\columncolor{mygray}}l}{37.927 $\pm$ 0.397}\\
\hline
ConvLSTM(h,f,s) &1.329 $\pm$ 0.016 &1.489 $\pm$ 0.022 &1.634 $\pm$ 0.022  &38.688 $\pm$ 0.644\\
GCNLSTM(h,f,s)  &1.405 $\pm$ 0.042 &1.569 $\pm$ 0.047 &1.714 $\pm$ 0.048  &41.041 $\pm$ 1.259\\
\midrule
\midrule
\multicolumn{5}{c}{Addition of covariates (c)}     \\
\hline
\multicolumn{1}{>{\columncolor{mygray}}l}{MHSTN-E(h,f,s,c)}  &\multicolumn{1}{>{\columncolor{mygray}}l}{1.310 $\pm$ 0.014}  &1.469 $\pm$ 0.012 &1.591 $\pm$ 0.012   &37.933 $\pm$ 0.398 \\
MHSTN-E(h,f,s,c*) &1.410 $\pm$ 0.021  &1.594 $\pm$ 0.027  &1.724 $\pm$ 0.022   &42.123 $\pm$ 0.721 \\ 
\hline
ConvLSTM(h,f,s,c) &1.392 $\pm$ 0.027  &1.508 $\pm$ 0.024 &1.650 $\pm$ 0.020  &39.218 $\pm$ 0.651\\
ConvLSTM(h,f,s,c*) &1.465 $\pm$ 0.042  &1.636 $\pm$ 0.043 &1.788 $\pm$ 0.048 &42.756 $\pm$ 1.351 \\
DUQ(h,f,c)        &1.712 $\pm$ 0.046  &1.992 $\pm$ 0.007 &2.634 $\pm$ 0.006 &66.911 $\pm$ 0.971  \\
DUQ(h,f,c*)        &1.728 $\pm$ 0.019  &2.030 $\pm$ 0.028 &2.669 $\pm$ 0.029 &71.338 $\pm$ 3.273 \\
\bottomrule
\end{tabular}
\end{table}

\begin{table}[htbp]
\caption{Results of the incremental evaluation with normalized metrics}
\label{tab:res-inc-norm}
\centering
\renewcommand\tabcolsep{5.0pt}
\begin{tabular}{p{2.6cm}p{1.75cm}p{1.75cm}p{1.75cm}p{1.9cm}}
\toprule
              & \multicolumn{4}{c}{Target wind variables}  \\
              \cline{2-5}
Models              & $v$ (RRSE)    &$vx$ (RRSE)   & $vy$ (RRSE)    & $\theta$ (NAMAE) \\  
\midrule
\multicolumn{5}{c}{Benchmarks}     \\
\hline
Persistence    &1.280  &1.316  &1.285   &0.383  \\
NWP            &0.863  &0.781  &0.688   &0.223   \\
\midrule
\midrule
\multicolumn{5}{c}{Ingestion of historical(h) and/or future(f) sequences}     \\
\hline
LSTM(h)        &0.969 $\pm$ 0.017   &0.985 $\pm$ 0.007   &0.985 $\pm$ 0.011   &0.344 $\pm$ 0.006 \\
LSTM(f)        &0.857 $\pm$ 0.049   &0.772 $\pm$ 0.017   &0.712 $\pm$ 0.019   &0.226 $\pm$ 0.005 \\
LSTM(h,f)      &0.848 $\pm$ 0.036   &0.785 $\pm$ 0.018   &0.726 $\pm$ 0.021   &0.230 $\pm$ 0.006 \\
MLP(h)         &0.983 $\pm$ 0.018   &0.989 $\pm$ 0.017   &0.997 $\pm$ 0.015   &0.359 $\pm$ 0.008 \\
MLP(f)         &0.775 $\pm$ 0.011   &0.734 $\pm$ 0.010   &0.662 $\pm$ 0.008   &0.219 $\pm$ 0.003 \\
MLP(h,f)       &0.785 $\pm$ 0.014   &0.766 $\pm$ 0.015   &0.686 $\pm$ 0.012   &0.229 $\pm$ 0.005 \\
\hline
\multicolumn{1}{>{\columncolor{mygray}}l}{MHSTN-T(h,f)} 
              &0.763 $\pm$ 0.011   &0.729 $\pm$ 0.013 &0.648 $\pm$ 0.009   &0.214 $\pm$ 0.004\\
\hline
GBRT(f)         &0.786 $\pm$ 0.012  &0.761 $\pm$ 0.015 &0.686 $\pm$ 0.011 &0.220 $\pm$ 0.004  \\
GBRT(h,f)        &0.783 $\pm$ 0.014  &0.753 $\pm$ 0.015 &0.672 $\pm$ 0.010 &0.215 $\pm$ 0.004  \\
SVR(f)         &0.790 $\pm$ 0.005  &0.769 $\pm$ 0.004 &0.675 $\pm$ 0.003 &0.216 $\pm$ 0.001  \\
SVR(h,f)        &0.794 $\pm$ 0.005 &0.796 $\pm$ 0.004 &0.697 $\pm$  0.003  &0.224 $\pm$ 0.002 \\
DUQ(h,f)        &0.929 $\pm$ 0.004 &0.957 $\pm$ 0.003 &1.057 $\pm$ 0.002   &0.367 $\pm$  0.005\\
\midrule
\midrule
\multicolumn{5}{c}{Addition of spatial information (s)}     \\
\hline
\multicolumn{1}{>{\columncolor{mygray}}l}{MHSTN-S(h,f,s)}
               &0.757 $\pm$ 0.009 &0.719 $\pm$ 0.008 &0.645 $\pm$ 0.009  &0.211 $\pm$ 0.003\\
\multicolumn{1}{>{\columncolor{mygray}}l}{MHSTN-E(h,f,s)}
               &0.749 $\pm$ 0.007 
               &\multicolumn{1}{>{\columncolor{mygray}}l}{0.719 $\pm$ 0.007}  
               &\multicolumn{1}{>{\columncolor{mygray}}l}{0.640 $\pm$ 0.005} 
               & \multicolumn{1}{>{\columncolor{mygray}}l}{0.211 $\pm$ 0.002}\\
\hline
ConvLSTM(h,f,s) &0.754 $\pm$ 0.009 &0.730 $\pm$ 0.011 &0.658 $\pm$ 0.009  &0.215 $\pm$ 0.004\\
GCNLSTM(h,f,s)  &0.794 $\pm$ 0.023 &0.768 $\pm$ 0.023 &0.689 $\pm$ 0.019  &0.228 $\pm$ 0.007\\
\midrule
\midrule
\multicolumn{5}{c}{Addition of covariates (c)}     \\
\hline
\multicolumn{1}{>{\columncolor{mygray}}l}{MHSTN-E(h,f,s,c)}  
              &\multicolumn{1}{>{\columncolor{mygray}}l}{0.740 $\pm$ 0.008}  
              &0.720 $\pm$ 0.006 &0.641 $\pm$ 0.005   
              &\multicolumn{1}{>{\columncolor{mygray}}l}{0.211 $\pm$ 0.002} \\
MHSTN-E(h,f,s,c*) &0.794 $\pm$ 0.012  &0.779 $\pm$ 0.012  &0.698 $\pm$ 0.010   &0.234 $\pm$ 0.004 \\ 
\hline
ConvLSTM(h,f,s,c) &0.783 $\pm$ 0.015  &0.740 $\pm$ 0.012 &0.664 $\pm$ 0.008  &0.218 $\pm$ 0.004\\
ConvLSTM(h,f,s,c*) &0.823 $\pm$ 0.023  &0.801 $\pm$ 0.021 &0.724 $\pm$ 0.021 &0.238 $\pm$ 0.008 \\
DUQ(h,f,c)        &0.950 $\pm$ 0.024  &0.960 $\pm$ 0.003 &1.058 $\pm$ 0.003 &0.372 $\pm$ 0.005  \\
DUQ(h,f,c*)        &0.961 $\pm$ 0.011  &0.979 $\pm$ 0.013 &1.073 $\pm$ 0.012 &0.396 $\pm$ 0.018 \\
\bottomrule
\end{tabular}
\end{table}

\subsubsection{Results of prediction accuracy (Tab.~\ref{tab:res-roll}, Tab.~\ref{tab:res-roll-norm}, Tab.~\ref{tab:res-inc} and Tab.~\ref{tab:res-inc-norm}).}
First of all, under both evaluation strategies, the results of the two benchmarks, i.e., Persistence and NWP, are the same because they do not have a training process. The rest are statistical models learning from training data. It can be observed that these statistical models consistently perform better on incremental evaluation than rolling evaluation. Such a phenomenon illustrates the common sense that more training data usually helps to improve statistical models.
Besides, we observe that Persistence is consistently much worse than NWP. It demonstrates that the prediction tasks are highly challenging, and NWP can achieve a relatively low error. The performance difference between variables also suggests an increase in the predictive difficulty of $v, vx, vy$ and $\theta$. 

Second, comparing models that ingest historical(h) and/or future(f) sequences, we have following discoveries:
\begin{enumerate}
    \item Future information is critical to attain accurate predictions. Specifically, models that ingest future sequence are usually better than those that do not. 
    \item LSTM and MLP are apt to model historical and future information, respectively. Specifically, LSTM(h) is better than MLP(h) on all conditions. MLP(f) as the best competitor is much better than LSTM(f).  
    \item Our temporal module, MHSTN-T(h,f), is usually the best. It implies that the temporal module can hybridize LSTM and MLP networks in an effective manner to assimilate historical and future data. Specifically, under rolling evaluation, MHSTN-T(h,f) achieves the best on variables of $vx$, $vy$ and $\theta$ and is competitive with MLP(f) on $v$. Under incremental evaluation, MHSTN-T(h,f) is significantly the best on all variables. The results show that MHSTN-T(h,f) can get more profit from the increase of data than other models. 
    \item Conventional ML models, i.e., GBRT and SVR, are worse than neural networks, i.e., MLP(f) and MHSTN-T(h,f), all the time. 
\end{enumerate}

Third, comparing models that leverage spatial information (s), we have following discoveries: 
\begin{enumerate}
    \item MHSTN-S(h,f,s) consistently outperforms both MHSTN-T(h,f) and other competitors. It means that MHSTN's spatial module is effective to mine spatial dependences inherent in multi-station data
    \item MHSTN-E(h,f,s) is usually the best and reduce the error of MHSTN-S(h,f,s) further. It suggests that MHSTN's ensemble module can balance temporal and spatial information to produce more accurate predictions.
\end{enumerate}

Finally, feeding covariates into advanced models, we have following discoveries:
\begin{enumerate}
    \item The best competitor to leverage spatiotemporal data, i.e., ConvLSTM(h,f,s), significantly get worse when considering covariates, no matter selected covariates (c) or all covariates (c*). 
    \item Compared with MHSTN-E(h,f,s), MHSTN-E(h,f,s,c) gets better on partial variables. Meanwhile, the decreases of MHSTN on the worse variables are negligible and much less than the ConvLSTM's. These results indicate that MHSTN is more robust to the additional variances.
    \item Taking all available covariates into consideration always results in bad models, e.g., MHSTN-E(h,f,s,c*). Therefore, the covariate selection module is needed in real-world applications where domain knowledge are usually absent. Further improvements can be made by exploring a more effective manner to leverage the selected covariates instead of directly feeding them into models, but we leave this for future work.
\end{enumerate}

In addition, we observe that the lately advanced DNN weather prediction model, DUQ \citep{wang2019deep}, obtains bad results on all conditions and it is even worse than NWP by a large margin.  We speculate the possible reasons are two folds.  
The first reason may be the difference between scenarios. DUQ was designed on the hypothesis that \textit{"Each day and each station are independent."} However, in our scenario, as illustrated in Fig.\ref{fig:corr-obs}(c), dense stations in a local field are highly correlated with each other. Besides, DUQ targeted to the variables of wind speed, temperature and humidity that have been demonstrated to be more stable than the vector of wind speed, ($vx$, $vy$), and wind direction, $\theta$, in the above experimental results and existing studies \citep{grover2015a,masseran2013fitting}.
Second, the primitive DUQ was just evaluated on a really small dataset that just considered specific 9 days' data. This way did not consider seasonal diversification and may produce an overfitting model.

In summary, under both evaluation strategies, MHSTN achieves the best results. Especially, under the incremental evaluation, MHSTN significantly reduces the prediction errors of NWP on $v, vx, vy$ and $\theta$ by 13.59\%, 7.04\%, 6.58\% and 5.48\%, respectively. In terms of normalized evaluation metrics, the reductions on $v, vx, vy$ and $\theta$ are 14.25\%, 7.93\%, 6.97\% and 5.38\%, respectively. 
All the above results motivate that an effective unified model, i.e., MHSTN, is needed to leverage multi-source data that usually bring in more variances/uncertainties to confuse models.

\subsubsection{Results of covariate selection (Tab.~\ref{tab:cov}).}
We further inspect the proposed covariate selection module.
For a target variable, the importance values of covariates are computed and saved as a vector on each station.
For simplicity, we use the average of the importance vectors over respective stations as the final importance vector.
Tab. \ref{tab:cov} reports the results. For the historical data, it can be observed that a target variable is most correlated to its own history. The model picks up covariates with importance values greater than a given threshold of $0.2$ (marked in gray font). Similarly, for the future data, we can see that except for the self-future, all covariates are generally uncorrelated with the target variable. Thus, the model just considers the self-future (marked in gray font) as input.   

\begin{table}[htbp]
\caption{Importance values of the covariates regarding to their respective target variables.}
\label{tab:cov}
\centering
\begin{tabular}{l|ccc|ccc}
\toprule
 & \multicolumn{6}{c}{Target variables}  \\
 \cline{2-7}
 & \multicolumn{3}{c}{on historical data}  & \multicolumn{3}{c}{on future data} \\
Covariates & $v$ & $vx$ & $vy$ & $v$ & $vx$ & $vy$\\  
\midrule
$v$       & \multicolumn{1}{>{\columncolor{mygray}}c}{0.9440} & \multicolumn{1}{>{\columncolor{mygray}}c}{0.2476}  & \multicolumn{1}{>{\columncolor{mygray}}c}{0.4946} & \multicolumn{1}{>{\columncolor{mygray}}c}{1.0000} & 0.1192           & 0.0642 \\
$vx$      & \multicolumn{1}{>{\columncolor{mygray}}c}{0.3493} & \multicolumn{1}{>{\columncolor{mygray}}c}{0.9964}  & \multicolumn{1}{>{\columncolor{mygray}}c}{0.6254} & 0.2066          & \multicolumn{1}{>{\columncolor{mygray}}c}{1.0000}  & 0.0765\\
$vy$      & \multicolumn{1}{>{\columncolor{mygray}}c}{0.4479} & \multicolumn{1}{>{\columncolor{mygray}}c}{0.2351}  & \multicolumn{1}{>{\columncolor{mygray}}c}{0.7443} & 0.1092          & 0.1848           & \multicolumn{1}{>{\columncolor{mygray}}c}{1.0000}\\
$\theta$  & 0.1325          & 0.0721           & 0.0942 & 0.0203          & 0.0624           & 0.0618 \\
$rh$      & 0.0038          & 0.0050           & 0.0131 & 0.0001          & 0.0004           & 0.0005 \\
$slp$     & \multicolumn{1}{>{\columncolor{mygray}}c}{0.3713} & 0.0503           & 0.0547 & 0.0296          & 0.0128           & 0.0183\\
$tp$      & \multicolumn{1}{>{\columncolor{mygray}}c}{0.3101} & 0.1632           & 0.1837 & 0.0444          & 0.0093           & 0.0053\\
\bottomrule
\end{tabular}
\end{table}

\begin{figure}[htbp]
\centering
\includegraphics[width=0.8\textwidth]{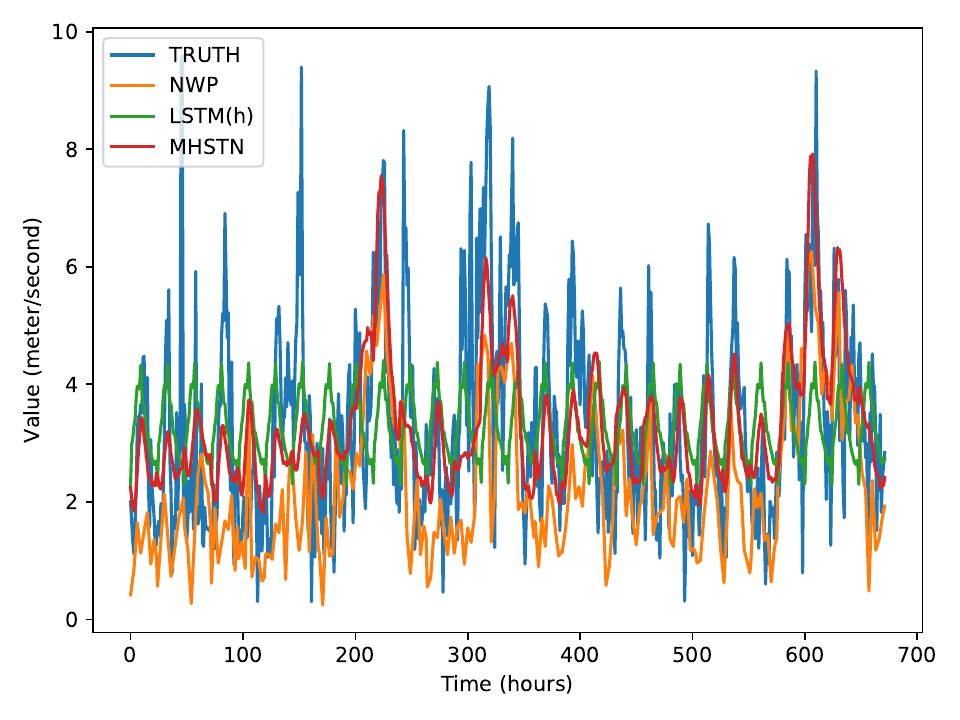}
\caption{The prediction results for $v$ at a station.}
\label{fig:vis-pred-v}
\end{figure}

\begin{figure}[htbp]
\centering
\includegraphics[width=0.8\textwidth]{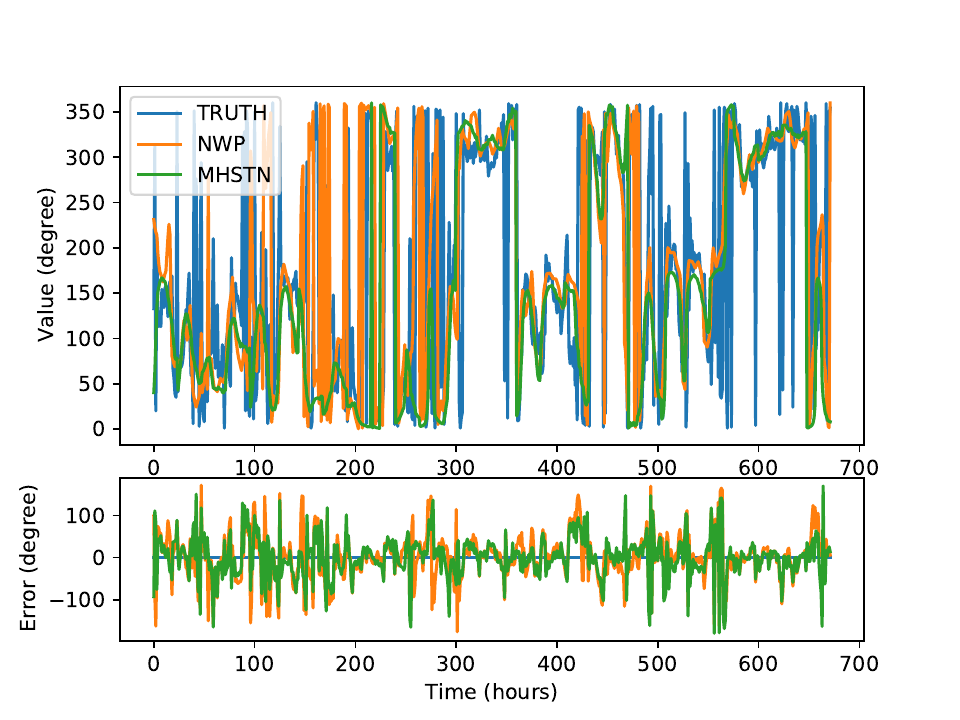}
\caption{The prediction results for $\theta$ at a station.}
\label{fig:vis-pred-dir}
\end{figure}

\subsubsection{Results of visualization (Fig.~\ref{fig:vis-pred-v} and Fig.~\ref{fig:vis-pred-dir}).}

For readability, we do not include all competitors that have been elaborated above. 
Fig.~\ref{fig:vis-pred-v} visualizes the prediction results of $v$ at a station where the testing set corresponds to the last split in the incremental evaluation. We can observe that LSTM(h) fails to capture the truth trajectory and the forecasts produced by this under-fitting model gather around the mean line. 
In contrast, by assimilating NWP data, MHSTN is able to align the future seasonality and events to generate sharp spiky forecasts. Further, the forecasts of MHSTN that injects additional local observation data looks better than that of NWP.
Fig.~\ref{fig:vis-pred-dir} visualizes the forecasts of $\theta$ in which we add a subplot to show the errors with regard to $\delta(\hat{\theta_{t}}, \theta_{t})$ defined in Eq.~\eqref{eq:mae-dir}. We can observe that the errors of NWP are greater than that of MHSTN on most timestamps. 

\subsubsection{Results of computation time.}
The applied NWP service is daily updated. To demonstrate MHSTN is efficient enough to make timely predictions, we report its computation time over the last split in the incremental evaluation that involves all data. To model a target variable (i.e., $v$) over all nine stations, the time costs for all modules are as follows. The training of the temporal module, the spatial module, the ensemble module, and the covariate selection module take 64.71 seconds, 51.74 seconds, 5.73 seconds, and 1.75 seconds, respectively. In the case of inference, the framework takes 0.21 seconds to produce predictions for one day or 24 hours.

\section{Conclusion}
\label{sec:con}
In this paper, we developed a unified Seq2Seq deep learning framework, MHSTN, for fine-grained multi-horizon wind prediction. MHSTN captures varying characteristics inherent in spatiotemporal weather data and has the capabilities of (1) efficiently predicting both wind speed and wind direction, (2) effectively fusing locally historical and globally estimated future information, (3) uniformly leveraging complicated correlations (including auto-correlation, cross-correlation and spatial correlation), and (4) simultaneously producing accurate multi-horizon predictions at a fine granularity.
We constructed a dataset using real-world data from one of the busiest international airport in China and conducted comprehensive experiments. The results demonstrated that synergy of MHSTN components enables it to outperform state-of-the-art competitors by a significant margin.
In the future, we are going to explore more advanced deep learning techniques to optimize each component of MHSTN and investigate the data assimilation problem of employing more data sources to increase the prediction performance. In the long run, we believe machine learning based techniques should be systematically integrated with traditional numerical frameworks to address varying trade-offs of short-/long-term and low-/high-resolution wind predictions.


%
%

\bibliographystyle{spbasic}      
\bibliography{reference}  


\end{document}